%% file: main.tex
\documentclass[runningheads]{llncs}

\usepackage{eccv}

\usepackage{eccvabbrv}

\usepackage{graphicx}
\usepackage{booktabs}

\usepackage[accsupp]{axessibility}  

\input{preamble}

\usepackage{hyperref}

\usepackage{orcidlink}

\begin{document}

\title{TRAIL: Trajectory-Aware Visual Place Recognition against Unordered Databases}

\titlerunning{TRAIL}

\author{Dominik A. Kloepfer\inst{1}\thanks{Work partially done during an internship at Helsing.}\orcidlink{0000-0002-1133-6023} \and
Patrick Wenzel\inst{2}\orcidlink{0000-0002-1846-6701}}

\authorrunning{D.A.~Kloepfer and P.~Wenzel}

\institute{Visual Geometry Group (VGG), University of Oxford \and
Helsing}

\maketitle
\input{sec/0_abstract}    
\input{sec/1_intro}
\input{sec/2_related_work}
\input{sec/3_method}
\input{sec/4_experiments}

\input{sec/5_conclusion}

\section*{Acknowledgements}
Dominik A.~Kloepfer gratefully acknowledges support from the EPSRC (VisualAI, EP/T028572/1).

%
%
\bibliographystyle{splncs04}
\bibliography{main}

\input{sec/X_suppl}

\end{document}

%% file: preamble.tex
\usepackage{enumitem}
\usepackage{wrapfig}
\usepackage{mathtools}
\usepackage{algorithm}
\usepackage{algorithmic}

\usepackage{svg}
\usepackage{tikz}
\usetikzlibrary{positioning}

\DeclareMathOperator*{\argmin}{arg\,min}
\DeclareMathOperator*{\argmax}{arg\,max}

\let\titleold\title
\renewcommand{\title}[1]{\titleold{#1}\newcommand{\thetitle}{#1}}

\usepackage{multibib}
\newcites{suppl}{References}

%% file: sec/0_abstract.tex
\begin{figure}
    \centering
    \vspace{-1.5em}
    \includegraphics[width=0.70\textwidth]{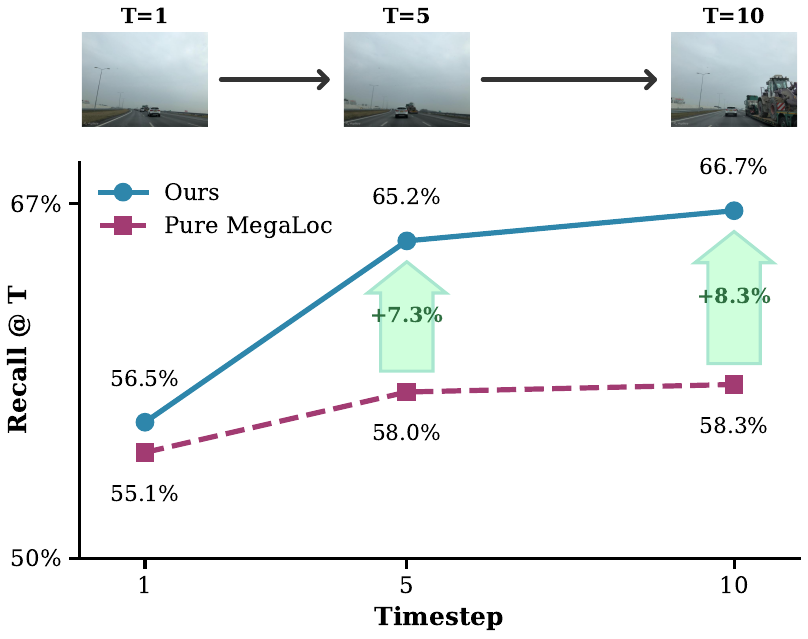}
    \caption{Incorporating information from previous images in a sequence boosts Visual Place Recognition (VPR) performance in challenging settings. We compare the top-1 Recall of the state-of-the-art VPR algorithm MegaLoc~\cite{berton2025megaloc} with our system, which makes use of previous images in the sequence (see \cref{sec:experiments} for full results).}
    \label{fig:splash}
    \vspace{-2.5em}
\end{figure}

\begin{abstract}
Modern Visual Place Recognition (VPR) methods excel on standard benchmarks yet remain brittle in feature-poor environments. By treating each query image in isolation, they discard the sequential context in any real trajectory. We formalize a task that exploits this context: given a query sequence, localize the \emph{final} image against an \emph{unordered} reference database---which, unlike sequence-to-sequence methods, requires no sequential structure in the database. We propose TRAIL (\textbf{TR}ajectory-\textbf{A}ware \textbf{I}mage \textbf{L}ocalization), a principled framework based on Conditional Random Fields (CRF) that combines learned functions for visual similarity and for camera-motion consistency, refining a distribution over candidate references as each query arrives. A lightweight post-processing layer atop any pre-trained VPR backbone, TRAIL improves a state-of-the-art baseline by up to 8.3 percentage points on our primary benchmark, transfers to unseen datasets without retraining, and delivers its largest gains where visual cues are scarce.
\end{abstract}

%% file: sec/1_intro.tex
\section{Introduction}
\label{sec:intro}

Visual Place Recognition (VPR), the task of matching query images against a reference database to determine camera location, has seen remarkable progress in recent years. Leveraging large-scale datasets and powerful pre-trained visual encoders such as DINOv2~\cite{oquab2023dinov2}, modern methods achieve impressive performance on established benchmarks~\cite{keetha2023anyloc, izquierdo2024salad, ali2024boq,berton2025megaloc}.

Yet, this progress masks a fundamental brittleness. In visually challenging scenarios such as rural landscapes, forests, or uniform highway environments, performance degrades sharply due to a lack of distinctive visual cues. Recent work has further exposed fragility in seemingly robust systems, revealing sensitivity to minor variations in evaluation parameters like retrieval distance thresholds~\cite{izquierdo2024cliquemining}.

This brittleness is particularly troubling given VPR's role in many important applications: camera relocalization, loop closure detection in SLAM, and hierarchical localization systems where VPR provides the initial coarse estimate.

We observe that standard VPR formulations exacerbate this brittleness through an artificial constraint: treating each query image in isolation. This single-image paradigm, while convenient for benchmarking, discards critical information available in any real navigation scenario. Cameras follow continuous trajectories through space, and each new observation arrives with strong geometric priors from preceding frames. Such information dramatically constrains the search space, but the standard VPR formulation is unable to leverage it.

Prior sequence-based VPR methods~\cite{garg2021seqnet, mereu2022sequencedescriptors2, berton2023jist, zhao2024sequencedescriptors, li2025casevpr} operate in a \emph{sequence-to-sequence} paradigm requiring the reference database to be organized as ordered sequences covering the same routes as potential queries. This imposes a significant structural constraint: it demands exhaustive coverage of all possible trajectories through the environment. In practice, however, reference imagery is typically collected opportunistically, from mapping vehicles, crowd-sourced platforms, or satellite data, yielding unordered sets of geo-tagged images rather than structured sequences.

In this paper, we formalize a task that addresses these limitations: given a sequence of query images, the goal is to localize the \emph{final} query image against an \emph{unordered} reference database. 
This setting captures the practical scenario of an agent that needs to determine its current position, exploiting the context from its recent trajectory without imposing any structural requirements on the database. 
We develop \emph{TRAIL} (\textbf{TR}ajectory-\textbf{A}ware \textbf{I}mage \textbf{L}ocalization), a principled probabilistic framework for this task, leveraging Bayesian filtering to maintain and update a probability distribution over reference locations as new query images arrive.
Crucially, this framework operates as a post-processing step over existing VPR methods, requiring no architectural modifications and minimal computational overhead.
 
Specifically, our contributions are as follows:

\begin{enumerate}[label=(\alph*)]
    \item We formalize a Sequential Visual Place Recognition task that operates on unordered reference databases and targets localization of the final query image, clearly distinguishing it from the sequence-to-sequence paradigm adopted by prior work (\cref{subsec:task_def}).
    \item We introduce \emph{TRAIL}, a probabilistic framework for incorporating sequential context into place recognition, designed to maximize retrieval accuracy while imposing minimal computational overhead (\cref{subsec:method_inference}, \cref{subsec:modeling_choices}, \cref{subsec:method_training}).
    \item We empirically validate our proposed framework and demonstrate its effectiveness on challenging datasets (\cref{sec:experiments}).
\end{enumerate}

By exploiting the temporal structure of navigation sequences, our method achieves notable improvements in both robustness and accuracy. Its simplicity and compatibility with existing systems make it well-suited for integration into real-world localization pipelines.

%% file: sec/2_related_work.tex
\section{Related Work}\label{sec:related_work}

\paragraph{Visual Place Recognition.}
Visual Place Recognition (VPR) has evolved from handcrafted local features and global descriptors~\cite{vlad2013, cummins2008fab, jegou2010aggregating, sunderhauf2011briefgist} to deep-learning techniques that brought remarkable performance gains, with architectures such as NetVLAD~\cite{arandjelovic2016netvlad}, Generalized Mean Pooling~\cite{radenovic2018gem}, and MixVPR~\cite{ali2023mixvpr} marking key milestones.
More recently, large pre-trained foundation models like DINOv2~\cite{oquab2023dinov2} have further boosted performance by acting as local feature extractors, with methods such as AnyLoc~\cite{keetha2023anyloc}, SALAD~\cite{izquierdo2024salad}, and Bag-of-Queries~\cite{ali2024boq} used to aggregate these features into global image descriptors. MegaLoc~\cite{berton2025megaloc} is a recent model that builds on the SALAD architecture and achieves state-of-the-art performance by combining training data from multiple datasets.

These advances have led to the saturation of several previously challenging benchmarks. However, challenges persist; for example, Izquierdo and Civera~\cite{izquierdo2024cliquemining} have illustrated the heavy reliance of recall performance on the distance threshold that determines correct retrievals. 

\paragraph{Sequential Visual Place Recognition.}
Sequence-based localization originates from loop closure detection, where SeqSLAM~\cite{milford2012seqslam} and its variants~\cite{siam2017fastseqslam, bai2018cnnseqslam} match query sequences against previously visited locations. This \emph{sequence-to-sequence} paradigm has since been extended to general VPR; SeqNet~\cite{garg2021seqnet} learns descriptors for sequence-based hierarchical retrieval, Mereu~\etal~\cite{mereu2022sequencedescriptors2} aggregate single-image descriptors into sequence-level representations, JIST~\cite{berton2023jist} jointly trains image and sequence descriptors, Zhao~\etal~\cite{zhao2024sequencedescriptors} apply spatio-temporal attention to construct sequence descriptors, and CaseVPR~\cite{li2025casevpr} performs correlation-aware alignment between query and database sequences. Closely related, Lynen~\etal~\cite{lynen2014placeless} match continuous query image streams against previously recorded trajectories, and SeqMatchNet~\cite{garg2022seqmatchnet} learns single-image descriptors that are explicitly shaped by a sequence-matching metric, collapsing to standard single-image retrieval when no ordered references are available.
Despite CaseVPR framing its task as \emph{sequence-to-frame}, its retrieval stage, too, operates over database sequences. All of these methods thus require the reference database to be organized as ordered sequences, which demands exhaustive route coverage of the environment.

The setting most closely related to ours is the \emph{seq2im} task described by Warburg~\etal~\cite{mapillary_sls}, which pairs query sequences with an unordered reference database. However, their formulation considers a retrieval correct if it is near \emph{any} image in the query sequence, rather than targeting a specific one. We formalize a stricter variant that reflects the practical localization objective: determining the position of the \emph{final} query image, and develop \emph{TRAIL}, a CRF-based framework, to address it.

\paragraph{Probabilistic Models for Sequence Localization.}
Some methods employ Hidden Markov Models (HMM) for sequence localization, such as the work by Rudi\'c~\etal~\cite{rudic2020indoorhmm} in indoor environments, and Gui~\etal~\cite{gui2024lidarhmm} who used LiDAR scans for unmanned ground vehicle localization in enclosed spaces. These approaches typically use discretized camera locations as hidden states, creating large state spaces that pose challenges in expansive environments. In contrast, we estimate directly the probability that database images are close to the true camera location, offering a novel approach to efficiently managing large environments.

Other approaches such as Cadena~\etal~\cite{cadena2012robustplacerecognition} use a formulation based on Conditional Random Fields (CRF), which we, too, use in our theoretical treatment (see \cref{subsec:method_inference} and Appendix~\cref{sec:appendix_derivation}). However, existing works do not address the problem of visual place recognition with a very large database but instead focus on loop closure in SLAM pipelines, which have a significantly smaller set of reference images. Most directly related, Xu~\etal~\cite{xu2021topometric} perform probabilistic topometric localization with a discrete Bayes filter whose forward recurrence coincides with the one we derive (Appendix~\cref{sec:appendix_derivation}). Their setting differs substantially, however, assuming an ordered reference map with per-edge odometry; transferring it to our unordered databases is non-trivial and forces simplifications that strip away much of its odometry-driven motion model, so that the resulting baseline fails to consistently improve over plain single-image retrieval (Appendix~\cref{sec:appendix_odometry_crf}).

More broadly, sequence and trajectory priors are widely used in the robotics-localization literature for loop-closure detection and verification in SLAM, with recent examples such as ROVER~\cite{yu2026rover} exploiting trajectory priors to reject false loop closures in repetitive environments. These methods, however, operate under far milder appearance variation than VPR (generally no day-night or season shifts) and assume the existence of \emph{loops} that revisit previously mapped locations; they therefore do not transfer straightforwardly to our setting, in which queries follow open trajectories with no guaranteed revisits and visual appearance can change drastically between query and reference.

%% file: sec/3_method.tex
\section{TRAIL: Sequential VPR with Conditional Random Fields}\label{sec:framework}

\subsection{Task Definition}\label{subsec:task_def}

Similarly to the standard Visual Place Recognition task, our goal is to localize an image with respect to a database set of $N$ \emph{unordered} reference images with associated locations, $\mathcal{R} = \{(r_i, x_i), i=1\dots N\}$, where $r_i \in \mathbb{R}^{3 \times H \times W}$ is an RGB-image of dimensions $H\times W$ and $x_i \in \mathbb{R}^3$ is a position coordinate.

In contrast to Visual Place Recognition, each query is not just a single query image, but a query \textit{sequence} of length $T$, so that $Q = (q_1, \dots, q_i, \dots, q_T) \in \mathcal{Q}_T$ with $q_i \in \mathbb{R}^{3 \times H \times W}$ images typically taken in order by an agent traversing the environment and $\mathcal{Q}_T$ the set of all query sequences of length $T$. Each image $q_i$ in the query sequence also has an associated location $y_i \in \mathbb{R}^3$, the agent's location when collecting this query image.

Given such a query sequence, a Sequential-VPR algorithm retrieves a single reference image, so we can formally write $a: \mathcal{Q} \rightarrow \mathcal{R}$. A retrieval $(r_i, x_i)$ is defined to be a \textit{correct} retrieval if $|x_i - y_T| \leq \delta$ with $\delta > 0$. In other words, the algorithm attempts to localize the \textit{final} query image in the query sequence by retrieving a reference image that was collected from a location within a certain threshold distance $\delta$ from the location of the query image.

This reflects the practical goal of localizing an agent \textit{now}, \ie, at the end of its trajectory.
We assume that reference images carry known positions (GPS or geo-tags), which is standard in VPR~\cite{berton2025megaloc,mapillary_sls}. At inference time, reference positions are used for the distance cutoff $\Delta$ (\cref{eq:transition_network}) and the aggregation function $\kappa$ (\cref{eq:scaled_sigmoid}).

We measure performance using a \textit{Recall@T} metric: the fraction of query sequences of length $T$ for which the retrieved image is correct. The standard VPR \textit{Recall@1} metric corresponds to the special case $T=1$. Note that $T$ denotes the sequence length, not the number of retrieved candidates as in the standard \textit{Recall@K} metric; the two coincide only for $T{=}K{=}1$.

\begin{figure*}
    \centering
    \includegraphics[width=\textwidth]{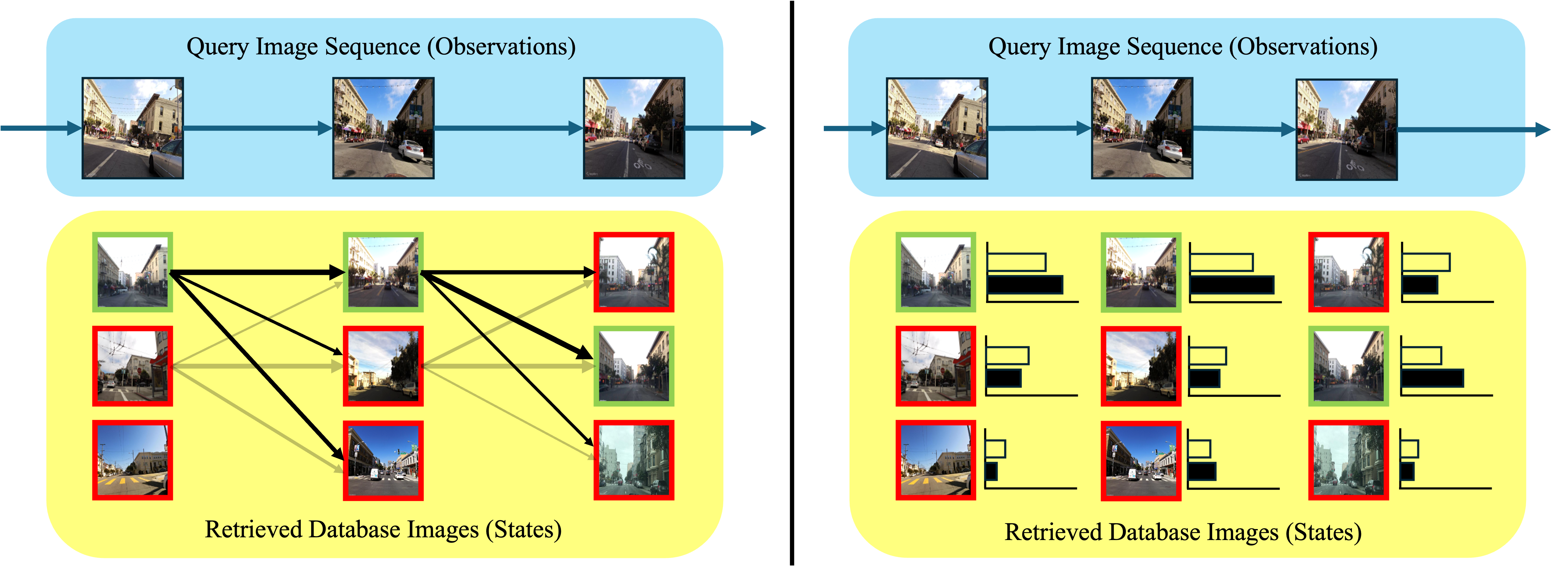}
    \caption{Diagram illustrating \emph{TRAIL}, our approach for Sequential Visual Place Recognition. For each query image $q_t$ in the sequence (top row), a pre-trained VPR model retrieves the top-$K$ database images, shown below in order of decreasing emission potential $\Psi_e$ (green border = correct retrieval, red = incorrect). The emission potential measures visual similarity between a query and a candidate reference image, while the transition potential $\Psi_{tr}$ (arrows, \textbf{left}) measures whether the spatial displacement between two candidate reference images is consistent with the observed camera motion between consecutive queries; thicker arrows indicate higher transition potentials. \textbf{Right}: Compared to using only emission potentials (top bars, not filled), fusing emission and transition potentials via the CRF recurrence (\cref{eq:recurrence}) yields substantial accuracy gains (bottom bars, filled; see \cref{subsec:results}).}
    \label{fig:seqvpr_diagram}
\end{figure*}

We now introduce \emph{TRAIL}, which leverages a pre-trained VPR algorithm to compute the probability that each database image is a correct retrieval for the final query image in a sequence, returning the location of the highest-probability reference image.

\subsection{Inference}\label{subsec:method_inference}

Our goal is to localize the final query image in the sequence, $q_T$, as reliably as possible, maximizing the \textit{Recall@T}. In other words, we want to return the reference image that has the highest probability of having been taken from a position within the threshold $\delta$ of the query image's position.

We proceed in two stages: First, we model $P(c_t^i | (q_1, \dots, q_t))$ where $c_t^i$ is a random variable with $c_t^i =1$ if the reference image $r_i$ is the retrieval image closest to query image $q_t$ and $c_t^i = 0$ otherwise.
Second, we use these probabilities to estimate $P(s_t^i | (q_1, \dots, q_t))$, where $s_t^i$ is a random variable with $s_t^i =1$ if the reference image $r_i$ is within the distance threshold $\delta$ of the query image $q_t$ and $s_t^i = 0$ otherwise.

In the appendix \cref{sec:appendix_derivation}, we show that under some light locality assumptions we can model this process using a Conditional Random Field (CRF)~\cite{lafferty2001crf} and derive the following recurrence relation for the first stage:

\begin{equation}
    P(c_t^i | Q) = \frac{\alpha_T(i)}{\sum_j \alpha_T(j)},
\end{equation}
where
\begin{align}\label{eq:recurrence}
    \alpha_1(i) &= \Psi_e(c_1^i, q_1) \\
    \alpha_t(i) &= \sum_j \Psi_{tr}(c_t^i, c_{t-1}^j, q_t, q_{t-1}) \Psi_e(c_t^i, q_t) \alpha_{t-1}(j) \nonumber.
\end{align}

$\Psi_e(c_t^i, q_t)$ and $\Psi_{tr}(c_t^i, c_{t-1}^j, q_t, q_{t-1})$ are two probability potentials. Intuitively, $\Psi_e$ measures the probability that the state where $c_t^i = 1$ would `emit' query image $q_t$, effectively measuring the similarity of the query image and a given retrieved image. $\Psi_{tr}$ in turn measures the probability that the state where $c_{t-1}^j = 1$ transitions to the state where $c_t^i = 1$, conditioned on the query images at $t-1$ and $t$.

Both of these probability potentials can be parameterized using neural networks $\phi_e(r_i, q_t) \approx \Psi_e(c_t^i, q_t)$ and $\phi_{tr}(r_i, r_j, q_t, q_{t-1}) \approx \Psi_{tr}(c_t^i, c_{t-1}^j, q_t, q_{t-1})$.

In the second stage, we use the computed $P(c_t^i | Q)$ to compute $P(s_t^i | Q)$, the probability of reference image $r_i$ being within the distance threshold of the query image $q_t$. 
We approximate 
\begin{equation}\label{eq:probability_aggregation}
    P(s_t^i | Q) = \sum_j P(s_t^i | Q, c_t^j) P(c_t^j | Q) \approx \sum_j P(s_t^i | c_t^j) P(c_t^j | Q).
\end{equation}
This approximation allows us to model $P(s_t^i | c_t^j) = \kappa(|x_i - x_j|)$ using some appropriate function $\kappa: \mathbb{R}^+ \rightarrow [0, 1]$.

The complete forward algorithm is given in \cref{algo:inference}. To obtain the predicted query image location at time step $t$, we return the location of the highest-probability reference image: $\hat{y}_t = x_{i^\star}$ where $i^\star = \argmax_j P(s_t^j | Q)$.

\begin{algorithm}
\caption{Forward Algorithm for Sequential VPR}
\label{algo:inference}
\begin{algorithmic}[1]

\REQUIRE $\phi_e, \phi_{tr}$ \COMMENT{neural networks}
\REQUIRE $\mathcal{R} = \{(r_i, x_i)\}$ \COMMENT{reference image database}
\REQUIRE $Q = (q_1, \dots, q_T)$ \COMMENT{query sequence}
\vspace{1ex}
\STATE $\alpha_1(i) \gets \phi_e(r_i, q_1) \forall r_i \in \mathcal{R}$
\vspace{1ex}
\FOR{$t = 2\dots T$}
    \STATE $\alpha_t(i) \gets \sum_j \phi_{tr}(r_i, r_j, q_t, q_{t-1}) \phi_e(r_i, q_t) \alpha_{t-1}(j)$
\ENDFOR
\vspace{1ex}
\FOR{$i: (r_i, x_i) \in \mathcal{R}$}
\STATE $P(c_T^i | Q) \gets \frac{\alpha_T(i)}{\sum_j \alpha_T(j)} $
\STATE $P(s_T^i | Q) \gets \sum_j \kappa(|x_i - x_j|) P(c_T^j | Q)$
\ENDFOR
\RETURN $P(s_T^i | Q),\; i=1\dots |\mathcal{R}|$
\end{algorithmic}
\end{algorithm}

\subsection{Network Architecture and Modeling Choices}\label{subsec:modeling_choices}

In principle, we need to compute $\alpha_t$ for all images in the retrieval database. The large size of the database, however, makes this intractable, so we set $\Psi_e(c_t^i, q_t) = 0 \;\forall i \notin v_K(q_t)$ and only compute $P(c_t^i | Q)$ for the top-$K$ reference images retrieved by a pre-trained VPR algorithm $v_K$. 

Unless $K$ is very large, there is a non-negligible probability that none of the top-$K$ retrievals are actually within the distance threshold $\delta$ of a given query image. To account for this case, we add a dustbin state $r_\text{dustbin}$ with $s_t^\text{dustbin} = 1 \text{ iff } s_t^i = 0 \;\forall i \in v_K(q_t)$. We found improved performance by setting $\Psi_{tr}(c_t^i, c_{t-1}^\text{dustbin}, q_t, q_{t-1}) = \Psi_e(c_t^i, q_t)$, which corresponds to the case where the algorithm ``lost track'' of the query location.

\subsubsection{Choice of $\kappa$.}
We require $\kappa(0) = 1$ and $\kappa(x) = 0$ for $x > \delta$, with smooth interpolation between these points. 
Intuitively, $\kappa$ models the distribution of the minimum localization error $|x_i - y_t|$ when $c_t^i = 1$, \ie, reference image $r_i$ is the best possible reference image; if this error was always zero and the location of the best possible reference image always coincided with the query image location, the optimal $\kappa$ would be $\kappa(x) = \textbf{1}(x \leq \delta)$ with $\textbf{1}$ as the indicator function.

We choose a scaled sigmoid:
\begin{equation}
    \kappa(x)= \mathbf{1}(x \leq \delta) \cdot \sigma(\frac{\gamma -x}{\tau}) / \sigma(\frac{\gamma}{\tau}) \label{eq:scaled_sigmoid}
\end{equation}
with the offset $\gamma$ fixed and the bandwidth $\tau$ learnable. The scaling factor $\sigma(\frac{\gamma}{\tau})$ ensures that $\kappa(0) = 1$, while the indicator function $\mathbf{1}$ ensures that $\kappa(x) = 0 \;\forall x > \delta$. This function consists of an initial flat section followed by a quick decay towards zero.
The boundary condition $\kappa(x){=}0$ for $x{>}\delta$ directly mirrors the evaluation criterion (a retrieval is correct iff it lies within $\delta$). Just as any retrieval method must define what counts as correct, $\kappa$ encodes the same notion into the probability model. Our ablation (\cref{sec:appendix_ablations}) shows that replacing the scaled sigmoid with a supergaussian yields similar performance, confirming that the framework is robust to the particular functional form of $\kappa$.

\subsubsection{Network Architectures.}
Finally, we also need to make some modeling choices for the network architectures of $\phi_e$ and $\phi_{tr}$. Since we already use a pre-trained visual place recognition algorithm $v_K$ to constrain the candidate reference images at each time step, it would be efficient to use intermediate results computed by that algorithm.
In our experiments, we use the standard form of visual place recognition algorithms that compute global image descriptors $d_\text{global} \in \mathbb{R}^{d_g}$ by first extracting local image features $d_\text{local} \in \mathbb{R}^{h \times w \times d_l}$ that are then aggregated into a single vector. We make the following choices: 

First, we notice that the task performed by $\phi_e$ is similar to pure Visual Place Recognition without sequential information, and so to the task that the global image descriptors are trained to do. We therefore set 
\begin{equation}
    \phi_e(r_i, q_t) = \text{MLP}(d_g(r_i) * d_g(q_t))\label{eq:emission_network}
\end{equation}
with $*$ corresponding to element-wise multiplication and $\text{MLP}$ denoting a small multi-layer perceptron, as a sort of `generalized cosine-similarity'.

To compute the probability potentials $\Psi_{tr}$, we first want to compute descriptors $d_{tr}$ for the movement between the reference images $r_j$ and $r_i$ and between the query images $q_{t-1}$ and $q_t$, and then assign a high probability potential if these transitions are similar. For the latter step, we again use a generalized cosine-similarity
\begin{equation}
    \phi_{tr}(r_i, r_j, q_t, q_{t-1}) = \text{MLP}(d_{tr}(r_i, r_j) * d_{tr}(q_t, q_{t-1})).
\end{equation}
We compute the transition descriptors $d_{tr}$ by computing the cosine-similarity between each pair of local image features and arranging them into a tensor of shape $\mathbb{R}^{h \times w \times (h\cdot w)}$, which we pass through a convolutional neural network:
\begin{equation}
    d_{tr}(a, b) = \text{CNN}(d_l(a) d_l(b)^\top).\label{eq:transition_descriptors}
\end{equation}

Finally, we notice that the probability potential $\Psi_{tr}$ should be zero for pairs of retrieval images that are very far apart, as a transition between those would be physically impossible. We therefore set $\phi_{tr} = 0$ when the positions $x_i, x_j$ of the retrieval images $r_i, r_j$ are too far apart. This leaves us with
\begin{equation}\label{eq:transition_network}
    \phi_{tr}(\dots) = \begin{cases}
        0 \text{ if } |x_i - x_j| > \Delta \\
        \text{MLP}(d_{tr}(r_i, r_j) * d_{tr}(q_t, q_{t-1})) \text{ otherwise}.
    \end{cases}
\end{equation}

\subsubsection{Intuition.}
Intuitively, the recurrence \cref{eq:recurrence} suppresses candidate reference images whose locations are inconsistent with earlier observations: if an earlier query image was confidently matched to database image $D$, for the current query image the transition potentials will down-weight candidate images at locations relative to $D$'s location that do not match the query images' trajectory. 

Even when no single image is easy to match (\eg, along a feature-poor highway), the transition potentials implicitly compare the relative camera motion inferred from reference image pairs with the motion observed in the query sequence, progressively constraining the set of plausible locations.

\subsection{Training}\label{subsec:method_training}

\subsubsection{End-to-End Training.}
The entire algorithm \cref{algo:inference} is differentiable, so given a pre-trained algorithm $v_K$ we train the entire system end-to-end by applying a binary cross-entropy loss on the computed $P(s_t^i | Q)$:

\begin{align} \label{eq:timestep_loss}
    \mathcal{L}_t = \sum_i &-\log(P(s_t^i | Q))\cdot\textbf{1}(|x_i - y_t| \leq \delta) \\
                        - &\log(1-P(s_t^i | Q))\cdot \textbf{1}(|x_i - y_t| > \delta). \nonumber
\end{align}

To improve training dynamics, we apply this loss at every timestep, not just $T$:

\begin{equation}
    L = \sum_t \mathcal{L}_t.
\end{equation}

In our implementation, we perform all calculations in log-space for improved numerical stability. The relevant derivations and equations can be found in the appendix in \cref{sec:appendix_logspace_derivation}.

\subsubsection{Pre-Training.}
End-to-end training is sped up and stabilized by first pre-training $\phi_e$ and $\phi_{tr}$ individually with cross-entropy losses.

For $\phi_e$, given query images with $K$ candidate retrievals from $v_K$, we train using:
\begin{equation}\label{eq:emission_loss}
    L_e = -\sum_i \log \frac{\exp\left(\phi(r_i^c, q_i)\right)}{\sum_k \exp(\phi(r_i^k, q_i))}
\end{equation}
where $r_i^c = \argmin_{r_i^k} |x_i^k - y_i|$ is the closest candidate to the query position.

For $\phi_{tr}$, we sample consecutive query pairs $(q_i^t, q_i^{t+1})$ and, for efficiency, train transitions \textit{from} the correct retrieval $r_i^{t,c}$ to $K$ candidates $r_i^{t+1,k}$. That is, for each pair of query images we also select $r_i^{t,c}$, the candidate retrieval with the position closest to $q_i^t$, and $K$ candidate retrievals $r_i^{t+1,k}$. We then compute $\phi_{tr}(r_i^{t, c}, r_i^{t+1,k}, q_i^t, q_i^{t+1})$ and again apply a cross-entropy loss:
\begin{equation}\label{eq:transition_loss}
    L_{tr} = -\sum_i \log \frac{\exp\left(\phi_{tr}(r_i^{t, c}, r_i^{t+1,c}, q_i^t, q_i^{t+1})\right)}{\sum_k\exp\left(\phi_{tr}(r_i^{t, c}, r_i^{t+1,k}, q_i^t, q_i^{t+1})\right)}.
\end{equation}

%% file: sec/4_experiments.tex
\section{Experiments}\label{sec:experiments}

\subsection{Datasets}\label{subsec:datasets}
\subsubsection{Mapillary Street-Level Sequences}
To train and evaluate TRAIL, we need datasets containing images with associated camera locations, organized into individual image sequences. A large and diverse dataset that fits these requirements is the \emph{Mapillary Street-Level Sequences (MSLS)}~\cite{mapillary_sls} dataset, and accordingly we use it as our main training dataset. 
This dataset contains 1.6 million images collected from short sequences of dashcam footage in 30 cities around the globe, providing a large and diverse set of images for training and evaluation.

Since individual sequences are split across the original `query' and `database' sets for each city, we merge them and construct a new database set for each sequence, against which localization is performed. For each sequence, the database set consists of all images from the same city that do not appear in the query sequence, and all results for MSLS cited in this paper refer to this setting.

To avoid degenerate sequences with little camera movement, we skip frames in a sequence that have a distance of less than $25.0$ meters from the previous frame.

\paragraph{Evaluation Protocol.}
Prior work \cite{keetha2023anyloc, izquierdo2024salad} holds out the images collected in San Francisco and Copenhagen for validation. However, we notice that these two cities are among the easiest in the entire dataset, with some of the highest \textit{Recall@1} performance for state-of-the-art Visual Place Recognition algorithms such as MegaLoc~\cite{berton2025megaloc}. We therefore instead hold out the images collected in Amsterdam and Boston for validation, which are more challenging.
Whereas MegaLoc has a \textit{Recall@1} performance of 89\% for images from Copenhagen and 84.5\% for images from San Francisco, the \textit{Recall@1} performance for images from Amsterdam is 54.0\% and 71.5\% for images from Boston.

We continue to use the distance threshold of $\delta = 25.0$ meters used in the literature \cite{keetha2023anyloc,izquierdo2024salad,berton2025megaloc}.

Two main factors account for the slightly worse \textit{Recall@1} performance on the Mapillary-SLS dataset in our setting compared to the performance reported on the validation split of the dataset elsewhere in the literature even for images from Copenhagen and San Francisco (\eg, Berton \etal~\cite{berton2025megaloc} reach $91.0\%$ \textit{Recall@1}). 
Firstly, we ensure that no images from the same sequence as a query image are included in the database. 
Secondly, we use \textit{all} other sequences to create the database instead of a more curated sub-set of images. 
While this leads to a higher theoretically possible localization performance due to a denser coverage of a given environment, it also increases the number of distractor images among which a VPR-algorithm must discriminate.

\subsubsection{Nordland}
Another dataset containing image sequences that can be used as query sequences in our setting is the \emph{Nordland} dataset~\cite{sunderhauf2013nordland}.
This dataset comprises videos of a train traversing a 728-kilometer route in Norway across four seasons. 

The largely rural landscape, with few distinctive features that remain consistent across seasons, makes this a challenging setting. While its lack of visual diversity makes it unsuitable for training, we evaluate our model on Nordland without fine-tuning, using weights trained on MSLS.

\paragraph{Evaluation Protocol.}
The standard test split of the Nordland dataset discards the sequential nature of the source data.
We therefore define the query sequences by uniformly sub-sampling 256 sequences of 10 frames each from the winter-video. 

As the database, we choose all the images of the summer-video, which is the season with the largest domain shift to the query season.

We filter out degenerate cases with little camera movement by skipping frames if they were not collected at least 10 meters from the previously sampled frame. 
Like for the MSLS dataset, we define $\delta = 25.0$ meters as the threshold for correct localization.

\subsubsection{4Seasons}
As a third dataset, we evaluate on the \emph{4Seasons}~\cite{wenzel2020fourseasons,wenzel2024fourseasons} dataset, which consists of gray-scale videos recorded using a sensor suite on a car following a number of different routes throughout different seasons.
Most of these routes lead through urban environments with a large number of visual features, which causes non-sequential VPR algorithms to saturate them. We therefore use the ``countryside'' loop, which is the most challenging route that also supplies ground-truth camera poses for each frame for evaluation. Like Nordland, this dataset is used for evaluation only, again using weights trained on MSLS.

\paragraph{Evaluation Protocol.}
We follow a similar protocol as for the Nordland dataset, extracting $252$ query sequences from the ``winter'' recording, which we localize against the ``summer'' recording that serves as the database. As before, we ensure that subsequent query frames are separated by at least $25.0$ meters and that database images are separated by at least $10.0$ meters, and use $\delta = 25.0$ meters as the threshold for correct localization. The comparatively small loop leaves a total of $672$ reference images.

\subsubsection{Other Datasets}
Few other datasets meet the requirement of sequential queries covering sufficiently large areas.
Standard VPR benchmarks such as Pittsburgh-30k~\cite{arandjelovic2016netvlad,gronat2013pittsburgh}, Tokyo 24/7~\cite{torii2015tokyo247}, and St.\ Lucia~\cite{warren2010stlucia} do not preserve the sequential structure of their source data in the standard evaluation splits, making them unsuitable for our task without non-trivial re-processing.
The Oxford RobotCar dataset~\cite{RobotCarDatasetIJRR} contains sequential information but covers a small, feature-rich loop where im2im methods are near-saturated.
Other domains such as off-road or aerial data are promising future directions where sequential priors may be especially valuable. We hope that clearly defining our task (\Cref{subsec:task_def}) and framework (\Cref{sec:framework}) will encourage future dataset collection and benchmark design in these domains.

\subsection{Architecture and Training}\label{subsec:arch_training}
We train on MSLS, holding out the Amsterdam and Boston images for evaluation and resize all images to squares with side-length $224$ pixels. 

As our pre-trained VPR model $v_K$ we use MegaLoc~\cite{berton2025megaloc}, which computes $8448$-dimensional global descriptors.
The emission network $\phi_e$ is a small MLP operating on the element-wise product of global descriptors (\cref{eq:emission_network}). The transition network $\phi_{tr}$ computes pairwise correlations between DINOv2~\cite{oquab2023dinov2} local features (ViT-14-Base, $768$-dim) and processes them with a ResNet-style CNN to obtain transition descriptors (\cref{eq:transition_descriptors}). We set the distance cutoff $\Delta = 75.0$ meters.

We keep MegaLoc frozen and only train $\phi_e$ and $\phi_{tr}$, first individually and then end-to-end using AdamW~\cite{loshchilov2018adamw} with RandAugment~\cite{cubuk2020randaugment} data augmentation (excluding geometric transforms, as they interfere with the learning of geometric transitions between images).

For pre-training, we use $K=10$ retrievals per query, with learning rate $10^{-3}$ and batch sizes of $B=56$ (emission) and $B=256$ (transition). For end-to-end training, we use batches of $6$ sequences of length $10$ with $K=10$ retrievals and learning rate $10^{-4}$.

All training is done on a single NVIDIA A100 GPU with 48GB memory. Full architecture details are provided in \cref{sec:appendix_arch_training}.

\paragraph{Computational and Memory Cost.}
TRAIL adds little overhead on top of the underlying VPR backbone. The DINOv2 local features used by $\phi_{tr}$ are produced by MegaLoc \emph{en route} to its global descriptor, so $\phi_{tr}$ requires no additional backbone pass, only the small CNN and MLP heads. Taking TRAIL (MegaLoc) as the unit of inference wall-clock per query, raw MegaLoc and the hand-crafted heuristics cost $0.2\times$ and FoL $0.6\times$, while the SelaVPR and PairVPR re-rankers cost $7.2\times$ and $9.3\times$ respectively. TRAIL is therefore substantially lighter than the re-rankers it matches or exceeds in accuracy (\cref{subsec:results}).
On the memory side, the cached local features for a single image occupy roughly $1.3\times$ the storage of the raw reference image ($5.3$, $14.8$, $14.6$, and $0.5$\,GB for MSLS-Amsterdam, MSLS-Boston, Nordland, and 4Seasons, respectively). Since only the $2K=20$ feature sets for the candidate retrievals of two consecutive frames are needed at once, these can be streamed from disk for arbitrarily large maps.

\subsection{Baselines}

We evaluate the performance of our approach against a number of baseline algorithms. We use the authors' code and pre-trained weights.

\subsubsection{Global Descriptor Algorithms.} 
MegaLoc~\cite{berton2025megaloc} and FoL~\cite{wang2025fol} are two recent, state-of-the-art Visual Place Recognition algorithms that compute global image descriptors and use these to retrieve matches from a database. We apply them to our setting by ignoring the sequential information and simply localizing the final image in the sequence. Note that while MegaLoc uses a ViT-B-based feature backbone, the authors of FoL only release weights for a ViT-L-based backbone. FoL therefore benefits from significantly larger model capacity, which should be kept in mind when interpreting its results.

\subsubsection{Sequence-Retrieval Algorithms.}
We also compare to methods trained to use sequences for Visual Place Recognition: SeqNet~\cite{garg2021seqnet}, JIST~\cite{berton2023jist}, and CaseVPR~\cite{li2025casevpr}. These methods assume that the database contains sequences of images, whereas our database consists of \emph{unordered} collections of images. We apply them by using the full query sequence descriptors to retrieve unordered database images, treating each as a single-image sequence.
We adapt CaseVPR's two stages as follows: we first retrieve the top-$K=5$ database images using the full query sequence descriptor, then choose among them the image with the highest similarity to the final query image's descriptor. Further details are given in \cref{subsec:appendix_seq_retrieval}.

\subsubsection{Re-Ranking Algorithms.} 
The third type of Visual Place Recognition Algorithm we evaluate are two recent re-ranking algorithms, SelaVPR~\cite{lu2024selavpr} and PairVPR~\cite{hausler2025pairvpr}. These im2im methods first retrieve $K=10$ candidate database images using global image descriptors, before then re-ranking these images using more expensive methods. Like for the Global-Descriptor-based algorithms, we apply these algorithms to our setting by ignoring sequential information and only localizing the final image in the query sequence.

\subsubsection{Hand-crafted Algorithms.} 
Finally, we compare to three hand-crafted methods that use the MegaLoc algorithm for computing image descriptors for retrieval.

The first, \emph{Maximum Similarity}, is briefly discussed in Warburg~\etal~\cite{mapillary_sls} as the +MIN approach for their seq2im task. Given a window size $W=2$, this algorithm computes the top-1 retrieved image for the final $W$ query images and returns the reference image that was retrieved with the highest similarity. 

The second, \emph{Majority Voting}, also appears in Warburg~\etal, under the name +MODE. Given a window size $W=3$, this algorithm uses the top-$K=10$ images retrieved for each of the final $W$ query images and returns the reference image that was retrieved most often among these $W\cdot K$ images.

Finally, \emph{Location Filtering} returns the database image that is most similar to the final image in the query sequence after filtering out any database image that was not collected within a distance threshold $d=75.0$ meters of at least one of the top-$K=10$ retrievals for the previous $W=2$ images. This corresponds roughly to the maximum distance between two correctly retrieved images when their respective query images are separated by approximately $25$ meters. Formal definitions for these hand-crafted heuristics can be found in \cref{subsec:appendix_handcrafted_baselines_definitions}.

\subsection{Results}\label{subsec:results}

 \begin{table}
  \centering
  \setlength{\tabcolsep}{4pt}
  \resizebox{\textwidth}{!}{%
  \begin{tabular}{@{}lccccccccc@{}}
    \toprule
    & \multicolumn{3}{c}{MSLS} & \multicolumn{3}{c}{Nordland} & \multicolumn{3}{c}{4Seasons} \\
    \cmidrule(lr){2-4} \cmidrule(lr){5-7} \cmidrule(lr){8-10}
    Algorithm & \emph{R@T=1} & \emph{R@T=5} & \emph{R@T=10} & \emph{R@T=1} & \emph{R@T=5} & \emph{R@T=10} & \emph{R@T=1} & \emph{R@T=5} & \emph{R@T=10}\\
    \midrule
    \multicolumn{10}{c}{\textit{Global Descriptor}} \\
    MegaLoc~\cite{berton2025megaloc} & 55.1 & 58.0 & 58.3 & 73.8 & 73.8 & 68.8 & 69.4 & 69.1 & 68.6 \\
    FoL (ViT-L)~\cite{wang2025fol} & 63.8 & 62.3 & 66.7 & 53.5 & 53.9 & 55.5 & 74.6 & 74.6 & 74.2 \\
    \midrule
    \multicolumn{10}{c}{\textit{Sequence-Retrieval}} \\
    SeqNet~\cite{garg2021seqnet} & 17.4 & 5.8 & 8.3 & 3.9 & 2.7 & 2.3 & 23.8 & 18.3 & 17.7 \\
    JIST~\cite{berton2023jist} & 17.4 & 11.6 & 16.7 & 10.9 & 1.6 & 0.4 & 34.5 & 15.1 & 6.5 \\
    CaseVPR~\cite{li2025casevpr} & 33.3 & 27.5 & 25.0 & 18.4 & 7.0 & 7.8 & 58.7 & 21.8 & 22.2 \\
    \midrule
    \multicolumn{10}{c}{\textit{Re-Ranking}} \\
    SelaVPR~\cite{lu2024selavpr} & 71.0 & \textbf{65.2} & \textbf{66.7} & 64.8 & 64.1 & 60.9 & \underline{77.0} & 76.6 & 76.2 \\
    PairVPR~\cite{hausler2025pairvpr} & 59.4 & 52.2 & 50.0 & 69.5 & 69.5 & 66.0 & \textbf{92.1} & \textbf{91.7} & \textbf{91.5} \\
    \midrule
    \multicolumn{10}{c}{\textit{Hand-Crafted}} \\
     Max. Similarity & 55.1 & 39.1 & 66.7 & 73.8 & 43.0 & 39.1 & 69.4 & 66.3 & 65.7 \\
     Majority Voting & 55.1 & 37.7 & 50.0 & 73.8 & 19.5 & 15.2 & 69.4 & 46.0 & 46.0 \\
     Loc. Filtering & 55.1 & 60.9 & 58.3 & 73.8 & \underline{79.7} & \underline{76.2} & 69.4 & 69.4 & 66.9 \\
    \midrule
    \multicolumn{10}{c}{\textit{Ours}} \\
    TRAIL (MegaLoc) & 56.5 & \textbf{65.2} & \textbf{66.7} & 75.3 & \textbf{81.3} & \textbf{78.1} & 71.0 & \underline{78.2} & \underline{79.0} \\
    \bottomrule
  \end{tabular}%
  }
  \caption{\textit{Recall@T} for $T=1, 5, 10$ on the Amsterdam and Boston subsets of Mapillary SLS~\cite{mapillary_sls}, on Nordland~\cite{sunderhauf2013nordland}, and on the 4Seasons~\cite{wenzel2020fourseasons,wenzel2024fourseasons} countryside loop (Nordland and 4Seasons: winter queries localized against a summer database). $T{=}1$ corresponds to standard single-image VPR. Best results in \textbf{bold}, second-best \underline{underlined}; FoL uses a larger ViT-L backbone than MegaLoc (ViT-B).}
  \label{tab:results}
\end{table}

Our main results are given in \cref{tab:results}. We report the \textit{Recall@T} metric for $T=1, 5, 10$ timesteps on the Amsterdam and Boston subsets of the Mapillary-SLS dataset, using all sequences that contain at least five images with successive query images separated by at least $25$ meters, on the Nordland dataset, localizing query-sequences drawn from the winter recording against the images drawn from the summer recording, and on the 4Seasons countryside loop under the analogous winter-to-summer protocol.

Whereas the improvements over vanilla MegaLoc are negligible for the first image in a sequence (when no sequential information is available), they are substantial for later frames.
On MSLS at $T{=}10$, SelaVPR reaches 66.7\%, tying with our method. However, SelaVPR applies per-frame re-ranking using cross-attention over local features, which is considerably more expensive than our small MLPs operating on $K{=}10$ pre-retrieved candidates. FoL, despite using a larger ViT-L backbone, does not outperform our ViT-B-based method at $T{\geq}5$ on MSLS and falls clearly behind on Nordland.

Sequence-retrieval methods that assume a database consisting of \emph{sequences} perform very poorly in our setting.
The core issue is that these methods treat all query images equally rather than focusing on localizing the final image: their sequence descriptors average across all frames, so when database ``sequences'' consist of a single image the descriptor distributions are mismatched.
TRAIL avoids this by explicitly modeling transitions and returning the location estimate for the final timestep. A more detailed analysis is given in \cref{subsec:appendix_seq_retrieval}.

For $T{=}1$, the handcrafted heuristics reduce to pure MegaLoc. Among them, only Location Filtering consistently improves over vanilla retrieval. This heuristic performs well on Nordland's predictable train trajectory but is less effective on the more varied driving data in MSLS, where learned transition modeling is more robust.

On 4Seasons, the same patterns hold: our method again improves substantially over the raw MegaLoc backbone as well as over the hand-crafted heuristics and SelaVPR, remaining effective despite the compounded domain shift from snowy winter queries to a summer database and from colour training data to grayscale test images, and despite never being trained on grayscale imagery. The notable exception is PairVPR, which performs exceptionally well on this dataset. We attribute this to the comparatively small reference set of the countryside loop, which leaves far fewer distractor images, combined with PairVPR being more robust to the season and colour domain shift than SelaVPR.

In \cref{subsec:appendix_additional_baselines} we report the performance of these baselines using different values for their hyper-parameters.

For an ablation analysis investigating the impact of some of the individual design choices for our method, refer to \cref{sec:appendix_ablations} in the supplement.

%% file: sec/5_conclusion.tex
\section{Conclusion}\label{sec:conclusion}
In this work, we presented \emph{TRAIL}, a novel framework for Sequential Visual Place Recognition that exploits temporal context across a sequence of query images. Specifically, TRAIL uses localization estimates from earlier images in the sequence to improve place recognition for the current query image within a large reference database.
This approach was empirically shown to improve the \textit{Recall@T} metric by up to $8.3$ percentage points on MSLS, $9.3$ on Nordland, and $10.4$ on 4Seasons, with minimal computational overhead.

Several directions for future work remain: (i) extending to domains beyond street-level driving, such as aerial or off-road imagery; (ii) exploring richer architectures for $\phi_e$ and $\phi_{tr}$, including fine-tuning the backbone $v_K$ itself; (iii) directly modeling the evolving distribution of the query image's position rather than classifying the closest reference image, which could yield smaller localization errors; and (iv) further reducing the framework's compute and memory footprint, for instance through a learned down-projection of the redundant DINOv2 local features, or by exploiting odometry data to drive lighter-weight transition potentials.

%% file: sec/X_suppl.tex

\clearpage
\setcounter{page}{1}
\setcounter{section}{0}
\renewcommand{\thesection}{\Alph{section}}

\begin{center}
    \Large\textbf{\thetitle}\\
    \vspace{0.5em}
    Supplementary Material\\
    \vspace{1em}
\end{center}

\section{Derivation of Recurrence Relation}\label{sec:appendix_derivation}
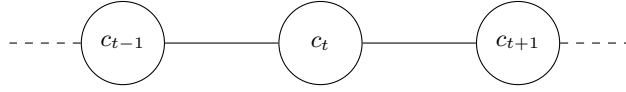
\begin{figure}[h!]
    \centering
    \begin{tikzpicture}[
         node_style/.style={
        circle, 
        draw, 
        text width=0.7cm,   
        align=center,       
        minimum size=1.1cm  
    }, 
    edge_style/.style={-}
    ]
        \node[node_style] (c1) {$c_{t-1}$};
        
        \node[node_style, right=1.5cm of c1] (c2) {$c_t$};
        \node[node_style, right=1.5cm of c2] (c3) {$c_{t+1}$};

        \draw[edge_style] (c1) -- (c2);
        \draw[edge_style] (c2) -- (c3);

        \draw[edge_style, dashed] (c1) -- ++(-1.5, 0);
        \draw[edge_style, dashed] (c3) -- ++(1.5, 0);
    \end{tikzpicture}
    \caption{Diagram of the undirected graph showing the conditional independence relations satisfied by the random variables $c_1, \dots, c_T$. After conditioning on the sequence of query images $Q$, each $c_t$ is independent of all $c_i$ s.t.\ $|i - t| > 1$, hence the set of random variables $\{c_t\}$ forms a Markov random field with respect to the graph.}
    \label{fig:markov_random_field}
\end{figure}
We are given a sequence of query images $Q = (q_1, \dots, q_i, \dots, q_T)$ with $q_i \in \mathbb{R}^{3\times H \times W}$ where each query image has an associated (but unknown) location $y_t \in \mathbb{R}^3$ and a set $\mathcal{R} = \{(r_i, x_i), i \in 1\dots N \}$ of reference images with associated locations, where $r_i \in \mathbb{R}^{3\times H \times W}$ and $x_i \in \mathbb{R}^3$. 

Our goal is to model $P(c_t | Q)$ for all $t=1\dots T$. Here we define $c_t = \argmin_i |x_i - y_t|$, so $c_t$ is the random variable that indicates which reference image $i$ was taken from the closest location to query image $q_t$. 

We first make the locality assumption that after conditioning on the observations $Q$, the random variables $c_t$ form a \emph{Markov Random Field} over the linear chain graph shown in \cref{fig:markov_random_field}. That means that each $c_t$ is conditionally independent of all other variables except for its neighbours $c_{t-1}$ and $c_{t+1}$, so $c_t \perp\!\!\!\!\perp \{c_i: |i - t| > 1\}$. 

By the Hammersley-Clifford Theorem \cite{hammersleyclifford1971}, we can write the conditional probability for each \emph{entire sequence} of closest states $C= (c_1, \dots, c_T)$ as a product over probability potentials (functions with a range $\mathbb{R}_{>0}$) of cliques of the graph. 

The cliques in the graph here (\cref{fig:markov_random_field}) are pairs of adjacent nodes and single vertices, so 

\begin{equation}
    P(C | Q) \propto \prod_{t=2}^T \Psi_{tr}(c_{t-1}, c_t, Q) \prod_{t=1}^T \Psi_e(c_t , Q)
\end{equation}

where $\Psi_{tr}, \Psi_e > 0$.

Intuitively, $\Psi_e$ should be large when $Q$ makes a particular value of $c_t$ likely, and $\Psi_{tr}$ should be large when the transition from $c_{t-1}$ to $c_t$ is plausible given $Q$.

This motivates the following further locality assumptions:
\begin{align}
\Psi_e(c_t, Q) &= \Psi_e(c_t, q_t) \\
\Psi_{tr}(c_{t-1}, c_t, Q) &= \Psi_{tr}(c_{t-1}, c_t, q_{t-1}, q_t) 
\end{align}
because those are the most influential query images for computing $\Psi_e$ and $\Psi_{tr}$, respectively.

We then get
\begin{align}
    P(C | Q) &\propto \Psi_e(c_1, Q) \prod_{t=2}^T \Psi_{tr}(c_{t-1}, c_t,Q) \Psi_e(c_t, Q) \\
            &\propto \Psi_e(c_1, q_1) \prod_{t=2}^T \Psi_{tr}(c_{t-1}, c_t, q_{t-1}, q_t) \Psi_e(c_t, q_t) \\
            &\propto S_T(C),
\end{align}
where we defined $S_T(C)$ as the unnormalised score for the state sequence $C$.

Let us now define $\alpha_t(i)$ as the unnormalised probability score of all state sequences $(c_1, \dots, c_{t-1}, i)$ that have length $t$ and end in state $c_t = i$. Then, trivially,
\begin{align}
    \alpha_1(i) = \Psi_e(c_1 = i, q_1),
\end{align}
and for other $t$ we have
\begin{align}
    \alpha_t(i)  =& \sum_{c_1, \dots, c_{t-1}} S_t(c_1, \dots, c_{t-1}, c_t=i) \\
                 =& \sum_{c_1, \dots, c_{t-1}}\Psi_e(c_1, q_1) \cdot \prod_{k=2}^{t-1} \Psi_{tr}(c_{k-1}, c_k, q_{k-1}, q_k) \Psi_e(c_k, q_k)\nonumber \\
                 &\;\;\;\;\;\;\;\;\;\;\;\;\;\cdot \Psi_{tr}(c_{t-1}, c_t=i, q_{t-1}, q_t) \Psi_e(c_t = i, q_t) \nonumber \\
                =& \Psi_e(c_t = i, q_t) \cdot \sum_j \Psi_{tr}(c_{t-1}=j, c_t=i, q_{t-1}, q_t)  \alpha_{t-1}(j), \nonumber 
\end{align}
which is the recurrence relation used in the main paper (\cref{eq:recurrence}). 

To calculate $P(c_T = i | Q)$, we have to normalise the $\alpha_t(i)$, so with the partition function $\mathcal{Z}_T(Q)$ we have
\begin{align}
    P(c_T=i | Q) &= \frac{\alpha_T(i)}{\mathcal{Z}_T(Q)} \propto \alpha_T(i) \\
                &= \frac{\alpha_T(i)}{\sum_j \alpha_T(j)},
\end{align}
as in the main paper.

\section{Derivation of Log-Space Equations}\label{sec:appendix_logspace_derivation}

For numerical stability, we implement the relevant equations in log-space. In this section, we derive the relevant log-space equations for the recurrence relation ($\cref{eq:recurrence}$), the probability aggregation (\cref{eq:probability_aggregation}), and the loss calculation (\cref{eq:timestep_loss}), as well as the log-space form of the scaled sigmoid function $\kappa$ (\cref{eq:scaled_sigmoid}). 

\subsection{Computing $P(c_t^i|Q)$ in Log-Space}

We start by taking the logarithm of both sides of the recurrence relation \cref{eq:recurrence}. 

\begin{align}
    \log \alpha_t(i) =& \log \left[\sum_j \Psi_{tr}(c_t^i, c_{t-1}^j, q_t, q_{t-1}) \Psi_e(c_t^i, q_t) \alpha_{t-1}(j) \right] \\
                    =& \log \Psi_e(c_t^i, q_t) + \log \left[\sum_j \Psi_{tr}(c_t^i, c_{t-1}^j, q_t, q_{t-1})\alpha_{t-1}(j) \right]. \nonumber
\end{align}
Defining $\phi_e = \log \Psi_e$ and $\phi_{tr} = \log \Psi_{tr}$, we get 
\begin{align}
    \log \alpha_t(i) =& \phi_e (c_t^i, q_t) + \text{logsumexp}_j [\phi_{tr}(c_t^i, c_{t-1}^j, q_t, q_{t-1}) + \log \alpha_{t-1}(j)].   
\end{align} 
Here $\text{logsumexp}$ can use the log-sum-exp trick to improve numerical stability. 

With this log-space recurrence relation, we then have
\begin{align}
    P(c_t^i | Q) &= \frac{\alpha_t(i)}{\sum_j \alpha_t(j)} \\
                &= \frac{\exp \log \alpha_t(i)}{\sum_j \exp \log \alpha_t(j)} \\
                &= \text{softmax}_i (\log \boldsymbol\alpha_t)
\end{align}
or
\begin{equation}
    \log P(c_t^i | Q) = \log \alpha_t(i) - \text{logsumexp}_j[\log \alpha_t(j)].
\end{equation}

\subsection{Computing $P(s_t^i | Q)$ in Log-Space}

To compute the binary cross-entropy loss on the aggregated probabilities $P(s_t^i | Q)$ (see \cref{eq:timestep_loss}), we need to compute both $\log P(s_t^i | Q)$ and $\log(1-P(s_t^i | Q))$.

Using $P(s_t^i | Q) = \sum_j \kappa(|x_i - x_j|) P(c_t^j|Q)$ from \cref{eq:probability_aggregation}, we find:
\begin{align}
    \log P(s_t^i | Q) = \text{logsumexp}_j \left[\log \kappa(|x_i - x_j|) + \log P(c_t^j | Q)\right]. 
\end{align}
To compute $1-P(s_t^i | Q)$, we can use the fact that $\sum_j P(c_t^j | Q) = 1$ and write
\begin{align}
    1-P(s_t^i | Q) &= 1 - \sum_j \kappa(|x_i - x_j|) P(c_t^j|Q) \\
                    &= \sum_j P(c_t^j | Q)- \sum_j \kappa(|x_i - x_j|) P(c_t^j|Q) \\
                    &= \sum_j (1 - \kappa(|x_i - x_j|)) P(c_t^j|Q).
\end{align}
Taking the logarithm of both sides then yields: 
\begin{align}\label{eq:log_1_minus_p}
    \log(1-P(s_t^i | Q)) = \text{logsumexp}_j [&\log(1 - \kappa(|x_i - x_j|)) + \log P(c_t^j | Q) ].
\end{align}

\subsection{Computing $\kappa$ in Log-Space}

To compute the overall loss, in log-space, the remaining piece is computing $\log \kappa(|x_i - x_j|)$ and $\log(1 - \kappa(|x_i - x_j|))$. 

In \cref{eq:scaled_sigmoid}, we defined
\begin{equation}
    \kappa(x)= \mathbf{1}(x \leq \delta) \cdot \sigma(\frac{\gamma -x}{\tau}) / \sigma(\frac{\gamma}{\tau})
\end{equation}
where $\sigma(x) = \exp(x)/(1 + \exp(x))$ is the sigmoid function and $\boldsymbol{1}$ is the indicator function, while $\gamma$ and $\tau$ are parameters.

We therefore have $\log \kappa(x) = -\infty$ for $x > \delta$, and for $x \leq \delta$
\begin{align}
    \log \kappa(x) =& \frac{\gamma-x}{\tau} - \log(1 + \exp(\frac{\gamma-x}{\tau})) - \frac{\gamma}{\tau} + \log(1 + \exp(\frac{\gamma}{\tau})) \\
                    =& \frac{x}{\tau} -\text{log1p}(\exp(\frac{\gamma-x}{\tau})) + \text{log1p}(\exp(\frac{\gamma}{\tau}))
\end{align}
where $\text{log1p}(x) = \log(1 + x)$, which has a numerically stable implementation for small $x$.

For $x> \delta$, $\log ( 1 - \kappa(x)) \rightarrow -\infty$, but since this term is exponentiated inside the $\text{logsumexp}$ in \cref{eq:log_1_minus_p}, we can handle it by simply masking out those terms.

Defining $A = \frac{ \gamma -x}{\tau}$ and $B = \frac{\gamma}{\tau}$ for notational simplicity, for $x \leq \delta$, we get
\begin{align}
    1-\kappa(x) =& 1 - \frac{\sigma(A)}{\sigma(B)} \\
                =& 1 -  \frac{\sigma(A)}{\sigma(B)} + \sigma(A) - \sigma(A) \\
                =& 1 - \sigma(A) + \left(1 - \frac{1}{\sigma(B)}\right) \sigma(A) \\
                =& \sigma(-A) + \frac{\sigma(B)-1}{\sigma(B)}\sigma(A) \\
                =& \sigma(-A) - \frac{\sigma(-B)}{\sigma(B)}\sigma(A),
\end{align}
where we used $\sigma(-x) = 1-\sigma(x)$. Now
\begin{align}
    \frac{\sigma(-B)}{\sigma(B)} =& \frac{1}{1 + \exp(B)} \frac{1 + \exp(-B)}{1} \\
                                =& \frac{1 + \exp(-B)}{1 + \exp(B)} \cdot \frac{\exp(B)}{\exp(B)} \\
                                =& \frac{\exp(B) + 1}{1 + \exp(B)} \cdot \exp(-B) \\
                                =& \exp(-B).
\end{align}
Therefore, 
\begin{align}
    1-\kappa(x) =& \sigma(-A) - \exp(-B)\sigma(A) \\
                =& \frac{1}{1 + \exp(A)} - \exp(-B)\frac{\exp(A)}{1 + \exp(A)} \\
                =& \frac{1 - \exp(A - B)}{1 + \exp(A)}.
\end{align}
Taking the logarithm on both sides and substituting back in for $A$ and $B$ finally yields
\begin{equation}
    \log(1 - \kappa(x)) = \text{log1p}\left[\exp\left(-\frac{x}{\tau}\right)\right] - \text{log1p}\left[\exp\left(\frac{\gamma - x}{\tau}\right)\right].
\end{equation}

\section{Architecture and Training Details}\label{sec:appendix_arch_training}

\paragraph{Emission Network.}
The MLP used in $\phi_e$ (\cref{eq:emission_network}) has input dimension $8448$ and a single hidden layer with dimension $64$, using LeakyReLU~\cite{maas2013leakyrelu} activations and dropout $0.1$.

\paragraph{Transition Network.}
The CNN computing transition descriptors $d_{tr}$ (\cref{eq:transition_descriptors}) is a ResNet~\cite{he2016resnet}-inspired network with three residual blocks. Each block contains two convolutional layers with ReLU~\cite{householder1941relu} activations and Batch-Normalization~\cite{ioffe2015batchnorm}, using hidden dimension $256$, kernel size $3$, and stride $1$. Mean-pooling and a linear projection produce $512$-dimensional descriptors. The MLP in $\phi_{tr}$ has two hidden layers of dimension $512$ and $256$, with LeakyReLU activations and dropout $0.1$. As local features we use the DINOv2~\cite{oquab2023dinov2} feature maps from MegaLoc (ViT-14-Base, patch size $14\times14$, feature dimension $768$).

\paragraph{Training.}
We train on MSLS, holding out Amsterdam and Boston for evaluation. Images are resized to $224\times224$ pixels with ImageNet normalisation, yielding local features of dimension $16 \times 16 \times 768$. We apply RandAugment~\cite{cubuk2020randaugment} with magnitude $9$, excluding translation, shear, and rotation augmentations as they interfere with learning geometric transitions.

For pre-training $\phi_e$: batches of $B=56$ query images with $K=10$ retrievals, AdamW~\cite{loshchilov2018adamw} with learning rate $10^{-3}$ and weight-decay $10^{-3}$.
For pre-training $\phi_{tr}$: batches of $B=256$ query pairs with $K=10$ retrievals, AdamW with learning rate $10^{-3}$ and weight-decay $10^{-4}$.
For end-to-end training: batches of $6$ sequences of length $10$ with $K=10$ retrievals, AdamW with learning rate $10^{-4}$ and weight-decay $10^{-3}$.

\section{Ablations}\label{sec:appendix_ablations}
\begin{table}
  \centering
    \small
  \begin{tabular}{@{}lcccccc@{}}
    \toprule
    & \multicolumn{3}{c}{MSLS} & \multicolumn{3}{c}{Nordland} \\
    \cmidrule(lr){2-4} \cmidrule(lr){5-7}
    Ablation & \emph{R@T=1} & \emph{R@T=5} & \emph{R@T=10} & \emph{R@T=1} & \emph{R@T=5} & \emph{R@T=10}\\
    \midrule
    No Pre-Training & 58.0 & \textbf{66.7} & 66.7 & \textbf{75.3} & 79.3 & 78.5 \\
     $\kappa = $\cref{eq:supergaussian} & 57.4 & \textbf{66.7} & 66.7 & 75.1 & 81.6 & 78.5 \\
     $\phi_{tr}$ uses $d_g$ & 56.5 & 65.2 & 66.7 & \textbf{75.3} & \textbf{83.2} & \textbf{80.5} \\
    $\phi_{tr} = 1$ & 56.5 & 60.9 & 62.7 & \textbf{75.3} & 76.2 & 73.4 \\
    TRAIL (FoL) & \textbf{60.9} & 59.4 & \textbf{75.0} & 55.1 & 61.3 & 53.1 \\
    \midrule
    TRAIL (MegaLoc) & 56.5 & 65.2 & 66.7 & \textbf{75.3} & 81.3 & 78.1 \\
    \bottomrule
  \end{tabular}
  \caption{Performance of Ablation Experiments on the MSLS and Nordland datasets. For detailed description of the experiments and a discussion of the results, please refer to \cref{sec:appendix_ablations}.}
  \label{tab:ablations}
\end{table}

To investigate the impact of some of our modeling choices on the system's performance, we perform a number of ablation experiments. Keeping all other training settings constant, we report the performance for a number of settings in \cref{tab:ablations}.



\subsubsection{No Pre-Training}
We experiment with training our model exclusively end-to-end, skipping the pre-training phase. While at early epochs this model's performance lags behind that of the pre-trained model, it quickly catches up and by the time the pre-trained model converges, achieves broadly similar performance.

\subsubsection{Alternative $\kappa$ Function.}
The choice of a scaled sigmoid as the functional form for the function $\kappa$ (\cref{eq:scaled_sigmoid}) is not unique. We also experiment with a similar function, a generalised Gaussian of the form 
\begin{equation}
    \kappa(x) = \exp(- (x / \tau)^\beta) \cdot \textbf{1}(x \leq \delta) \label{eq:supergaussian}
\end{equation}
where $\tau$ corresponds to the cut-off in the scaled-sigmoid and is set to $20.0$ meters and the shape parameter $\beta$ is learnable and initialised to $4.0$. 

This setting, too, achieves broadly similar performance to the model in the main paper. This is not surprising, as the functional shape of both equations is quite similar.

\subsubsection{$\phi_{tr}$ uses Global Descriptors.}
In the main paper, we use an architecture for the transition probability potentials that uses local features to compute descriptors $d_{tr}$ of the transition between a pair of images (\cref{eq:transition_descriptors}). 
In this ablation, we instead use the global image descriptors extracted by the pre-trained VPR algorithm to compute $d_{tr}$:
\begin{equation}
    d_{tr} = \text{MLP}(d_g(r_i) * d_g(q_t)).
\end{equation}
The Multi-layer Perceptron we use here has two hidden layers with $512$ dimensions each, and uses leaky-ReLU activations and a dropout of 0.1 during training.

The performance is slightly stronger than that of the full model on the Nordland dataset. 
MegaLoc~\cite{berton2025megaloc} uses the SALAD~\cite{izquierdo2024salad} architecture, which aggregates local features into clusters that are then concatenated. Such global descriptors may still retain information about the spatial arrangement of features in the image, which a sufficiently expressive network can exploit.
For the Nordland dataset especially, this information can be enough to identify the quite consistent camera movement between subsequent frames.

We also note that the number of parameters of this Multi-layer Perceptron ($4.8$ million) is larger than that for the CNN used in the main paper ($3$ million).

\subsubsection{Setting $\phi_{tr} = 1$.}
We also ablate the effect of explicitly integrating the information derived from the transitions between query images compared to the effect of explicitly modeling the probability $P(s_t^i|Q)$ by aggregating the probabilities $P(c_t^i|Q)$ using \cref{eq:probability_aggregation}.

We do so by setting $\phi_{tr}=1$ in our trained model, which results in the recurrence relation \cref{eq:recurrence} becoming $\alpha_t(i) \propto \phi_e(c_t^i, q_t)$. The proportionality constant cancels when computing $P(c_t^i | Q)$. 
The performance of this modification lags significantly behind the full model, though it still outperforms the pure MegaLoc baseline.

Standard VPR-Algorithms return the single reference image with the highest similarity with the query image, so do not take into account the similarity scores assigned to other reference images. 
By explicitly modeling the probability $P(s_t^i|Q)$, our approach allows a reference image with a slightly lower similarity score to be chosen ahead of an image with a higher similarity score, if there are other reference images with reasonably high scores at locations close to its location. This helps discard outlier similarity scores, where two unrelated places look similar from only a certain perspective or only in certain lighting conditions.

\subsubsection{Different Backbone.}
Finally, we evaluate our method trained using the FoL~\cite{wang2025fol} backbone (a recent VPR method using a ViT-L-based architecture), with the architecture and training recipe otherwise unchanged.

It, too, tends to improve on the raw FoL backbone, demonstrating that our framework is backbone-agnostic.
The gains are more inconsistent, though, which may stem from FoL's specific training recipe making it less suited to our specific choices for the emission and transition network architectures.

Our framework's \emph{principles} (\cref{subsec:method_inference}, \cref{subsec:method_training}, \cref{subsec:modeling_choices}, and Appendix \cref{sec:appendix_derivation}) remain applicable to any VPR backbone, however.

\subsection{Sequence Pre-Processing and Transition Cutoff}\label{subsec:appendix_protocol_ablations}

\begin{table}
  \centering
    \setlength{\tabcolsep}{4pt}
  \resizebox{\textwidth}{!}{%
  \begin{tabular}{@{}lccccccccc@{}}
    \toprule
    & \multicolumn{3}{c}{MSLS} & \multicolumn{3}{c}{Nordland} & \multicolumn{3}{c}{4Seasons} \\
    \cmidrule(lr){2-4} \cmidrule(lr){5-7} \cmidrule(lr){8-10}
    Setting & \emph{R@T=1} & \emph{R@T=5} & \emph{R@T=10} & \emph{R@T=1} & \emph{R@T=5} & \emph{R@T=10} & \emph{R@T=1} & \emph{R@T=5} & \emph{R@T=10}\\
    \midrule
    TRAIL (MegaLoc) & 56.5 & \textbf{65.2} & 66.7 & 75.3 & 81.3 & 78.1 & 71.0 & 78.2 & \textbf{79.0} \\
    No stationary-frame filter & 56.5 & 63.8 & 66.7 & 75.3 & 80.1 & 78.9 & 71.0 & 75.0 & 74.6 \\
    $\Delta = 50.0$\,m & 56.5 & 63.8 & 66.7 & 75.3 & \textbf{82.8} & \textbf{80.0} & 71.0 & \textbf{79.4} & \textbf{79.0} \\
    $\Delta = 100.0$\,m & 56.5 & 63.8 & 66.7 & 75.3 & 79.7 & 77.0 & 71.0 & 77.4 & 77.0 \\
    \bottomrule
  \end{tabular}%
  }
  \caption{Effect of removing the stationary-frame filter and of varying the transition cutoff $\Delta$. Performance is broadly stable across all three settings; best results per column in \textbf{bold} (columns in which all settings tie are left unmarked).}
  \label{tab:ablations_protocol}
\end{table}

We further study two choices in our data pre-processing and modeling pipeline that are independent of the trained networks: the filtering of near-stationary query frames and the value of the transition cutoff $\Delta$. For both, we re-run the \emph{same} trained TRAIL model without any retraining; the results are reported in \cref{tab:ablations_protocol}.

\subsubsection{Stationary and Variable-Speed Frames.}
To avoid degenerate sequences with little camera movement, we skip query frames that are less than $25.0$ meters from the previous frame (\cref{subsec:datasets}). We ablate this choice by re-running trained TRAIL on the \emph{unfiltered} sequences. Each \emph{Recall@}$T$ metric still refers to the same target query frame as before, but the previously-skipped intermediate frames now also flow through TRAIL. As \cref{tab:ablations_protocol} shows, performance remains broadly stable. In a deployment setting, such frames can also be removed cheaply using image similarity or coarse odometry, without recourse to ground-truth positions.

\subsubsection{Transition Cutoff $\Delta$.}
The distance cutoff $\Delta = 75.0$ meters, beyond which transition potentials are set to zero, is the geometric ceiling on the separation between two consecutive references: given a query spacing of $25$ meters and a localization threshold $\delta = 25.0$ meters, two correctly localized consecutive references can be at most $\delta + 25 + \delta = 75$ meters apart. Varying $\Delta \in \{50, 100\}$ meters yields similar results (\cref{tab:ablations_protocol}), confirming that the framework is insensitive to the precise value of this cutoff.

\section{Hand-Crafted and Sequential Baseline Details}

\subsection{Formal Definition}\label{subsec:appendix_handcrafted_baselines_definitions}

We now give the formal definitions of the three hand-crafted heuristics evaluated in the main paper in \cref{sec:experiments}. As we saw there, it is not trivial to create heuristics that outperform even the simplest baseline. 

For each of these algorithms, we assume the availability of a previously trained Visual Place Recognition algorithm, which, given a single query image, can retrieve the $K$ most similar images according to some algorithm-specific similarity metric. In the notation used above, such an algorithm is a function $v_K: \mathcal{Q}_1 \rightarrow \mathcal{R}^K$. Usually, such an algorithm computes a descriptor for the query image and all reference images and then retrieves the top-K database entries: $v_K(q) = \textrm{top-}K_{(r, x) \in \mathcal{R}} [s(d(q), d(r))]$, where $d: \mathbb{R}^{3 \times H \times W} \rightarrow \mathbb{R}^d$ is an image descriptor, and $s: \mathbb{R}^{2\times d} \rightarrow \mathbb{R}$ is a similarity function.

\subsubsection{Maximum Similarity.}
This baseline is briefly discussed in Warburg \etal~\cite{mapillary_sls} as an approach for their seq2im task, calling it +MIN. Given a window size $W \leq T$, this algorithm computes the top-1 retrieved image for the final $W$ query images and returns the reference image that was retrieved with the maximum similarity. Formally, this set of top-1 retrieved images for the final $W$ query images is $\mathcal{W}_1((q_1, \dots, q_T)) = \bigcup_{i=T-W+1}^T v_1(q_i)$ and so with 
\begin{equation}
    \hat{s}(r) = \max_{\substack{T-W <i \leq T\\ v_1(q_i)=r }} s[d(q_i), d(r)] \;\forall r \in\mathcal{W}_1
\end{equation}
the highest retrieval similarity for each retrieved image we have
\begin{equation}
    a_\textrm{max. similarity}(Q; v_K) = \argmax_{(r, x) \in \mathcal{W}_1(Q)} \hat{s}(r).\label{eq:maximum_similarity}
\end{equation}

\subsubsection{Majority Voting.}
This baseline is also briefly discussed in \cite{mapillary_sls}, under the name +MODE. Again, given a window size $W \leq T$, this algorithm computes the top-K images retrieved for each of the final $W$ query images and returns the reference image that was retrieved most often among these $W\cdot K$ images. With $\mathcal{W}_K((q_1, \dots, q_T)) = \bigcup_{i=T-W}^T v_K(q_i)$, we have formally
\begin{equation}
    a_\textrm{majority voting}(Q; v_K) = \textrm{mode}(\mathcal{W}_K(Q)).\label{eq:majority_voting}
\end{equation}

\subsubsection{Location Filtering.}
This baseline is built on the heuristic that correctly retrieved database images for the different images in the query sequence will have been collected from locations not too far apart from each other. It returns the database image that is most similar to the final query sequence image after filtering out any retrieved image that was not collected within a distance threshold $d$ of at least one of the top-K retrievals for the previous $W$ query images. Formally, 
\begin{equation}
a_\textrm{loc.filtering}(Q; v_K) = \argmax_{(r, x) \in \mathcal{F}_{K,W}} s[d(q_T), d(r)],\label{eq:location_filtering}
\end{equation}
where we define the set of filtered retrievals as 
\begin{equation*}
\mathcal{F}_{K,W} = \{(r, x) \in v_K(q_T) \textrm{ s.t. }\exists (s, y) \in \bigcup_{i=T-W}^{T-1} v_K(q_i) \textrm{ with } |x-y| \leq \Delta\}.
\end{equation*}

\subsection{Additional Baselines}\label{subsec:appendix_additional_baselines}

\begin{table}
  \centering
    \footnotesize
  \begin{tabular}{@{}lcccccc@{}}
    \toprule
    & \multicolumn{3}{c}{MSLS} & \multicolumn{3}{c}{Nordland} \\
    \cmidrule(lr){2-4} \cmidrule(lr){5-7}
    Setting & \emph{R@T=1} & \emph{R@T=5} & \emph{R@T=10} & \emph{R@T=1} & \emph{R@T=5} & \emph{R@T=10}\\
    \midrule
    Pure Retrieval (MegaLoc) & 55.1 & 58.0 & 58.3 & 73.8 & 73.8 & 68.8 \\
    \midrule
    \multicolumn{7}{c}{\textit{Maximum Similarity}} \\
    \cmidrule(lr){1-7}
    $W=2$ & 55.1 & 39.1 & 66.7 & 73.8 & 43.0 & 39.1 \\
    $W=3$ & 55.1 & 39.1 & 66.7 & 73.8 & 31.6 & 25.0 \\
    $W=5$ & 55.1 & 34.8 & 66.7 & 73.8 & 19.9 & 18.4 \\
    $W=10$ & 55.1 & 34.8 & 66.7 & 73.8 & 19.9 & 10.6 \\
    \midrule
    \multicolumn{7}{c}{\textit{Majority Voting}} \\
    \cmidrule(lr){1-7}
    $W=3, K=10$ & 55.1 & 37.7 & 50.0 & 73.8 & 19.5 & 15.2 \\
    $W=2, K=10$ & 55.1 & 30.4 & 58.3 & 73.8 & 38.3 & 34.8 \\
    $W=3, K=5$ & 55.1 & 42.0 & 58.3 & 73.8 & 22.3 & 16.0 \\
    $W=2, K=5$ & 55.1 & 37.7 & 58.3 & 73.8 & 39.1 & 36.3 \\
    $W=3, K=20$ & 55.1 & 36.2 & 41.7 & 73.8 & 21.5 & 16.0 \\
    $W=2, K=20$ & 55.1 & 29.0 & 58.3 & 73.8 & 40.2 & 36.7 \\
    \midrule
    \multicolumn{7}{c}{\textit{Location Filtering}} \\
    \cmidrule(lr){1-7}
    $W=2, K=10, \Delta=75.0$ & 55.1 & 60.9 & 58.3 & 73.8 & 79.7 & 76.2 \\
    $W=3, K=10, \Delta=75.0$ & 55.1 & 60.9 & 58.3 & 73.8 & 78.5 & 75.0 \\
    $W=3, K=10, \Delta=100.0$ & 55.1 & 60.9 & 58.3 & 73.8 & 78.5 & 74.2 \\
    $W=2, K=10, \Delta=50.0$ & 55.1 & 60.9 & 58.3 & 73.8 & 79.3 & 76.2 \\
    $W=3, K=10, \Delta=50.0$ & 55.1 & 60.9 & 58.3 & 73.8 & 78.1 & 75.0 \\
    $W=2, K=10, \Delta=25.0$ & 55.1 & 56.5 & 58.3 & 73.8 & 37.9 & 35.2 \\
    \midrule
    TRAIL (MegaLoc) & 56.5 & \textbf{65.2} & \textbf{66.7} & \textbf{75.3} & \textbf{81.3} & \textbf{78.1} \\
    \bottomrule
  \end{tabular}
  \caption{Additional Results for Hand-Crafted Baselines with Different Hyperparameters. We see that the performance of the baselines can sometimes be sensitive to the hyperparameter choices made, but does not reach the performance of our method.}
  \label{tab:additional_baselines}
\end{table}

We also evaluated the hand-crafted heuristics using a range of different hyper-parameters for the window size $W$, the number of candidate reference images $K$, and, in the case of the \emph{Location Filtering} baseline, the distance filtering threshold $\Delta$. The results of these experiments can be found in \cref{tab:additional_baselines}. 

The different hyperparameters fail to improve the overall performance, and the \emph{Majority Voting} baseline is quite sensitive to the choice of hyperparameter; larger window sizes in particular decrease its performance as there are more spurious votes by earlier query images for reference images that are too far away from the final query image.
 
For the \emph{Maximum Similarity} baseline, a higher window size primarily increases the likelihood that the highest-similarity retrieval will be for a previous time-step, and lie outside the distance threshold for correct localizations.

Similarly, the \emph{Location Filtering} baseline is quite insensitive to the exact hyperparameter choice, because its hard filtering approach generally filters out the most egregious similarity-score outliers, while harder examples are not filtered out for sensible choices of hyperparameters. On the Nordland dataset, though, we see that a filtering threshold of $\Delta = 25.0 \textrm{ meters}$ is too strict and significantly reduces performance.

These results underscore the need for careful thinking about how to best integrate sequence information into a visual place recognition system.

\subsection{Sequence-Retrieval Baselines}\label{subsec:appendix_seq_retrieval}

\begin{table}
  \centering
    \small
  \begin{tabular}{@{}lcccccc@{}}
    \toprule
    & \multicolumn{3}{c}{MSLS} & \multicolumn{3}{c}{Nordland} \\
    \cmidrule(lr){2-4} \cmidrule(lr){5-7}
    Algorithm & \emph{R@T=1} & \emph{R@T=5} & \emph{R@T=10} & \emph{R@T=1} & \emph{R@T=5} & \emph{R@T=10}\\
    \midrule
    SeqNet~\cite{garg2021seqnet} & 17.4 & 5.8 & 8.3 & 3.9 & 2.7 & 2.3 \\
    JIST~\cite{berton2023jist} & 17.4 & 11.6 & 16.7 & 10.9 & 1.6 & 0.4 \\
    CaseVPR~\cite{li2025casevpr} & 33.3 & 27.5 & 25.0 & 18.4 & 7.0 & 7.8 \\
    \midrule
    TRAIL (MegaLoc) & \textbf{56.5} & \textbf{65.2} & \textbf{66.7} & \textbf{75.3} & \textbf{81.3} & \textbf{78.1} \\
    \bottomrule
  \end{tabular}
  \caption{Sequence-retrieval baselines evaluated in our setting with an \emph{unordered} reference database. These methods are designed for databases organised into sequences and perform poorly when this assumption is violated.}
  \label{tab:seq_retrieval_baselines}
\end{table}

We evaluate three methods designed for sequence-to-sequence Visual Place Recognition---SeqNet~\cite{garg2021seqnet}, JIST~\cite{berton2023jist}, and CaseVPR~\cite{li2025casevpr}---in our setting, where the reference database consists of \emph{unordered} images rather than sequences (\cref{tab:seq_retrieval_baselines}).

These methods assume that the database is organised into sequences and compute fixed-length descriptors that aggregate information across all frames in a sequence. We apply them to our setting by using the full query sequence descriptors to retrieve the unordered database images, treating each database image as a single-frame ``sequence.'' For CaseVPR, which has two stages, we first retrieve the top-$K{=}5$ database images using the full query sequence descriptor, then choose among them the image with the highest similarity to the image descriptor of the final query image.

All three methods perform far below standard single-image retrieval. Two factors explain this:
\begin{enumerate}
    \item \textbf{Descriptor mismatch.} Sequence descriptors are trained to match against other sequence descriptors. When the database contains only single images, the descriptor distributions are mismatched---the query descriptor averages over $T$ frames while each database descriptor represents a single frame.
    \item \textbf{No focus on the final image.} These methods treat all frames in a sequence equally, producing a descriptor that represents the entire trajectory rather than privileging the final query image. Since earlier query images may be far from the final image's location, the aggregated descriptor is biased away from the target location. TRAIL avoids this by explicitly modeling transitions and returning the location estimate for the final timestep.
\end{enumerate}

These results confirm that sequence-to-sequence methods are not directly applicable to our task setting, motivating TRAIL, which works with unordered databases by design.

\section{Odometry CRF Baseline}\label{sec:appendix_odometry_crf}

\begin{table}
  \centering
  \setlength{\tabcolsep}{4pt}
  \resizebox{\textwidth}{!}{%
  \begin{tabular}{@{}lccccccccc@{}}
    \toprule
    & \multicolumn{3}{c}{MSLS} & \multicolumn{3}{c}{Nordland} & \multicolumn{3}{c}{4Seasons} \\
    \cmidrule(lr){2-4} \cmidrule(lr){5-7} \cmidrule(lr){8-10}
    Algorithm & \emph{R@T=1} & \emph{R@T=5} & \emph{R@T=10} & \emph{R@T=1} & \emph{R@T=5} & \emph{R@T=10} & \emph{R@T=1} & \emph{R@T=5} & \emph{R@T=10}\\
    \midrule
    MegaLoc~\cite{berton2025megaloc} & 55.1 & 58.0 & 58.3 & 73.8 & 73.8 & 68.8 & 69.4 & 69.1 & 68.6 \\
    Odometry CRF~\cite{xu2021topometric} & 47.8 & 58.0 & 41.7 & 73.4 & 76.6 & 75.0 & 69.1 & 69.8 & 52.5 \\
    TRAIL (MegaLoc) & \textbf{56.5} & \textbf{65.2} & \textbf{66.7} & \textbf{75.3} & \textbf{81.3} & \textbf{78.1} & \textbf{71.0} & \textbf{78.2} & \textbf{79.0} \\
    \bottomrule
  \end{tabular}%
  }
  \caption{Comparison against a hand-crafted odometry CRF adapted from Xu~\etal~\cite{xu2021topometric}. TRAIL outperforms both pure MegaLoc and the odometry CRF across all three datasets.}
  \label{tab:odometry_crf}
\end{table}

In addition to the baselines evaluated in the main paper, we compare against a hand-crafted odometry-based CRF adapted from Xu~\etal~\cite{xu2021topometric}, who perform probabilistic appearance-invariant topometric localization. The results are reported in \cref{tab:odometry_crf}.

Adapting their method to our setting is not straightforward: their motion model is built on an \emph{ordered} reference map whose adjacent places are joined by relative-pose edges, which we do not have. This in turn means our references carry no orientation, so the directional part of the query odometry cannot be used and only its magnitude survives; the simplifications this forces are derived in \cref{subsec:appendix_odometry_crf_derivation}. 
For odometry, we use real IMU integration on 4Seasons; on MSLS and Nordland, which do not provide inertial measurements, we derive odometry from the ground-truth positions corrupted with calibrated random-walk noise.

As \cref{tab:odometry_crf} shows, this adaptation trails pure MegaLoc on MSLS and 4Seasons, with the drop concentrated at large $T$, where the absence of heading information most hurts long-range disambiguation. Strengthening this baseline---for example by fusing visual odometry to recover a heading estimate---is a promising direction for future work.

We further note that our forward recurrence (\cref{eq:recurrence}) parallels Eq.~(10) of Xu~\etal~\cite{xu2021topometric}. Their transition potentials, however, condition on the query odometry, which violates the standard causality assumption of hidden Markov models and the textbook derivation of the forward algorithm; our derivation in \cref{sec:appendix_derivation} closes this gap.

\subsection{Adapting the TopometricLoc Motion Model}\label{subsec:appendix_odometry_crf_derivation}

We now derive the odometry CRF baseline in detail, adopting the notation of Xu~\etal~\cite{xu2021topometric} where convenient. Their TopometricLoc system is a discrete Bayes filter that maintains a belief over a set of reference places together with a single \emph{off-map} state $O$, which absorbs probability mass whenever the query leaves the mapped area. At each timestep the belief is updated by a \emph{measurement model}, scoring the appearance of the query against each reference, and a \emph{motion model}, a transition matrix $E_t$ that scores how consistent each candidate-to-candidate transition is with the query odometry. The two are combined through the standard hidden-Markov-model forward recursion.

This filter is structurally the same model as TRAIL (\cref{sec:appendix_derivation}). Its measurement model is an \emph{emission potential}, playing the role of our $\Psi_e$; its motion model is a \emph{transition potential}, playing the role of our $\Psi_{tr}$; its off-map state $O$ is our dustbin state; and its forward recursion is exactly ours. The substance of the adaptation lies in re-deriving these two potentials for an unordered reference set, which we do below.

Crucially, the TopometricLoc map is an \emph{ordered traverse}: each reference node carries an appearance descriptor, and \emph{adjacent} nodes are joined by edges that store a relative-pose estimate $\mu^r_{i\to j}$ obtained from mapping-time odometry. As we restrict the candidate set at each timestep to the top-$K$ retrievals of the underlying VPR model, the transition matrix is of size $(K{+}1)\times(K{+}1)$.

\paragraph{Emission Potential.}
The measurement model is the baseline's emission potential and transfers to our setting unchanged, as it requires only descriptors and is indifferent to whether the references are ordered. Following Xu~\etal, the emission factor for candidate $i$, the analogue of our $\Psi_e(c_t^i, q_t)$, is an exponential appearance kernel
\begin{equation}
    g_{t,i} \propto \exp\!\left(-\lambda \,\lVert d_g(q_t) - d_g(r_i)\rVert_2\right),
\end{equation}
where $d_g$ is the global descriptor of the underlying VPR model and the sharpness $\lambda$ is calibrated once, at $t{=}1$, from the spread (a pair of quantiles) of the distances between the first query descriptor and all map descriptors. The off-map state is assigned the emission of the $k$-th best match for some $k \gg 1$, yielding a likelihood that is consistently high but never optimal; the filter therefore only commits to a within-map location when the appearance matches are genuinely strong.

\paragraph{Transition Potential.}
The motion model is the baseline's transition potential, the analogue of our $\Psi_{tr}$. In the notation of Xu~\etal, the relative pose of the query at time $t$, given the hypothesis that it was at reference $i$ at time $t-1$, is modelled as a Gaussian $\mathcal{N}(\mu_t^q, \Sigma_t^q)$ parameterised by the \mbox{3-DoF} query odometry. The transition score $E_{t,ij}$ is the likelihood of this Gaussian evaluated on the \emph{mapped trajectory segment} $T_j$ around node $j$, a short curve interpolated between the relative poses of $j$'s neighbouring nodes, which reduces to a minimum squared Mahalanobis distance $d^2_{t,ij}$ between the predicted pose and $T_j$.

This construction relies on two ingredients the unordered setting denies us. First, the edges and per-edge relative poses $\mu^r_{i\to j}$ needed to interpolate $T_j$ do not exist when the references are an unordered set of points. Second, and more subtly, each reference carries a position but no \emph{orientation}: there is no local frame at node $i$ against which the \emph{direction} of a query displacement could be measured.
In Xu~\etal this frame is provided by the heading of the mapping traverse at each node, with which a directional query odometry is implicitly assumed to be aligned. We therefore make three generalisations.

\emph{(G1) A neighbourhood ball replaces the trajectory segment.} With unordered references there is no notion of a predecessor or successor of node $j$, and hence no segment $T_j$ to interpolate. We replace it with a 2D ball of radius $R$ around the candidate's position,
\begin{equation}
    \mathcal{N}(j) = \{y \in \mathbb{R}^2 : \lVert y - x_j\rVert \leq R\},
\end{equation}
and set $R = \delta$, the same neighbourhood radius that defines a correct localization in our evaluation. Lacking a local trajectory, the most we can assert is that a correct candidate lies somewhere within a metric neighbourhood of $x_j$.

\emph{(G2) Isotropic odometry covariance.} We take $\Sigma_t^q = \sigma^2 I$ with a constant scalar $\sigma$, so the Mahalanobis distance collapses to a scaled Euclidean distance.

\emph{(G3) Comparison of displacement magnitudes only.} Our odometry yields a 2D displacement between consecutive frames, but its \emph{direction} cannot be used. There is no per-reference heading to compare it against (above), and the absolute orientation of the query trajectory in the map frame is unobservable to a system localizing against an unordered database; supplying it would, for our simulated odometry, amount to feeding in ground-truth heading. 
We therefore retain only the rotation-invariant displacement \emph{magnitude} $s_t = \lVert o_t - o_{t-1}\rVert$ between the odometry-derived query positions $o_{t-1}, o_t$.
Writing the map separation between candidates as $\Delta d_{ij} = \lVert x_j - x_i\rVert$, the predicted query position given hypothesis $i$ then lies somewhere on the \emph{circle} of radius $s_t$ centred on $x_i$. The minimum distance from this circle to the ball $\mathcal{N}(j)$ is $\max\{0,\, |\Delta d_{ij} - s_t| - R\}$ (treating the radial offset as one-dimensional), which yields the squared cost
\begin{equation}
    d^2_{t,ij} = \frac{1}{\sigma^2}\Big(\max\{0,\; |\Delta d_{ij} - s_t| - R\}\Big)^2 .
\end{equation}
This discards all directional information, precisely the cue that most helps disambiguate distant places that happen to lie a similar travel distance away, and is reflected in the baseline's degradation at large $T$ (\cref{tab:odometry_crf}).

With $d^2_{t,ij}$ in hand, the transition matrix is assembled exactly as in Xu~\etal. The within-map block is a softmax over the negative half-costs, rescaled by the complementary off-map mass,
\begin{equation}
    \log E_{t,ij} = -\tfrac{1}{2} d^2_{t,ij} - \text{logsumexp}_{j'}\!\left(-\tfrac{1}{2} d^2_{t,ij'}\right) + \log\!\left(1 - E_{t,iO}\right),
\end{equation}
while the within-to-off-map probability grows with the cost of the best available transition through a chi-squared CDF,
\begin{equation}
    E_{t,iO} = \chi^2_\nu\!\left(\min_j d^2_{t,ij}\right),
\end{equation}
capped at a maximum per-step value. We set the degrees of freedom to $\nu = 2$, matching our planar motion, rather than the $\nu = 3$ used by Xu~\etal for \mbox{3-DoF} motion. Finally, transitions \emph{out of} the off-map state fall back to the emission factors, $\log E_{t,Oj} = \log g_{t,j}$.

\paragraph{Inference.}
Localization then proceeds by the same forward recursion and decoding as our own model. In log-space,
\begin{equation}
    \log \alpha_t(j) = \log g_{t,j} + \text{logsumexp}_i\!\left(\log E_{t,ij} + \log \alpha_{t-1}(i)\right),
\end{equation}
initialised with the emission factors $\log \alpha_1(j) = \log g_{1,j}$. This is exactly our forward recurrence (\cref{eq:recurrence}), with the emission factor $g_{t,j}$ in place of $\Psi_e$ and the transition factor $E_{t,ij}$ in place of $\Psi_{tr}$. As before, we normalise $\alpha_t$ with a softmax, aggregate the resulting state distribution over candidate locations with the same neighbourhood aggregation used throughout our framework (\cref{eq:probability_aggregation}), and return the highest-scoring candidate. The baseline thus differs from TRAIL only in its potentials---hand-crafted, odometry-driven factors in place of the learned $\phi_e$ and $\phi_{tr}$. The one conceptual gap is that conditioning the transition potential on the query odometry makes the recurrence depart from the textbook forward algorithm; our learned model avoids this by deriving appearance-only potentials from first principles in \cref{sec:appendix_derivation}.

\section{Qualitative Analysis}

\begin{figure*}[t]
\centering
\begin{subfigure}[b]{0.48\textwidth}
    \centering
    \includegraphics[width=\textwidth]{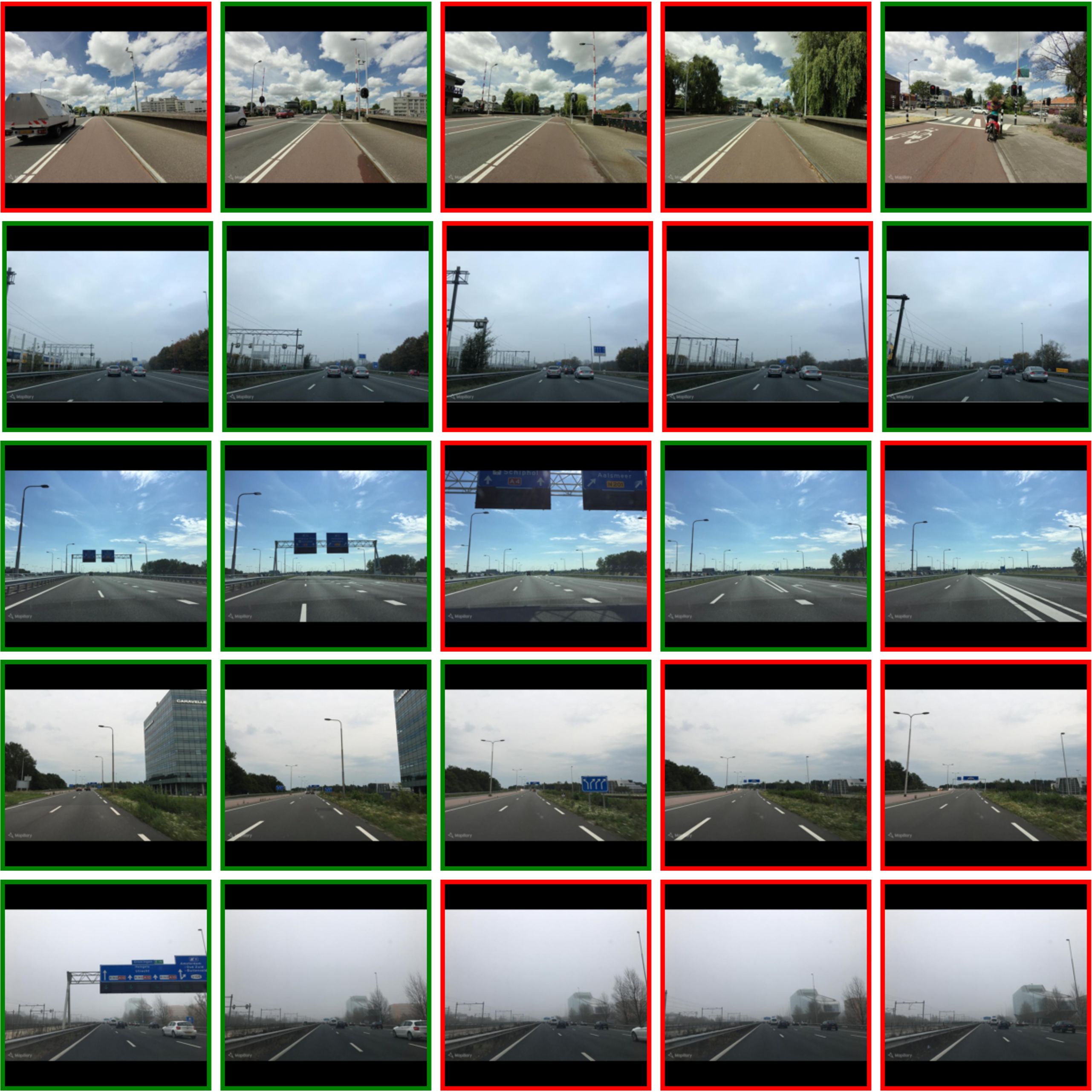}
    \caption{\emph{Pure Retrieval (MegaLoc)} Baseline}
    \label{sub_fig:qualitative_retrieval_only}
\end{subfigure}
\hfill
\begin{subfigure}[b]{0.48\textwidth}
    \centering
    \includegraphics[width=\textwidth]{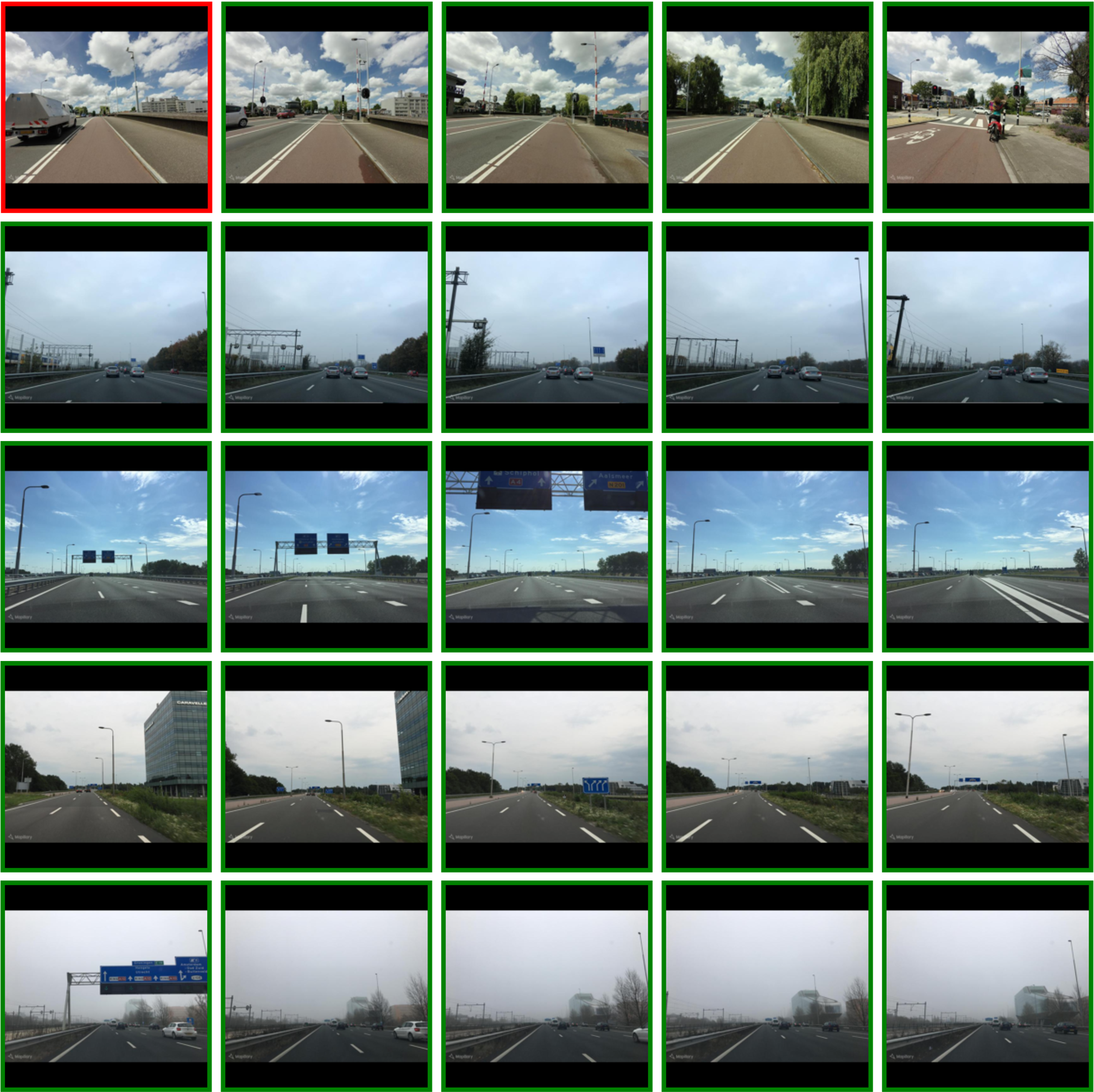}
    \caption{Our \emph{TRAIL (MegaLoc)}}
    \label{sub_fig:qualitative_seqvpr}
\end{subfigure}
\caption{Selected Validation Sequences where \emph{TRAIL} outperforms \emph{Pure Retrieval}. Feature-sparse rural and high-way regions as opposed to feature-rich urban regions are overrepresented compared to the full validation set. Green borders indicate correct, red borders incorrect localization. The performance of the \emph{Pure Retrieval} baseline is shown to the left in \cref{sub_fig:qualitative_retrieval_only}, the performance of our \emph{TRAIL} method to the right in \cref{sub_fig:qualitative_seqvpr}.}
\label{fig:qualitative_examples}
\end{figure*}

In this section, we give some qualitative examples of query sequences in the Mapillary dataset for which our \emph{TRAIL} method outperforms basic Visual Place Recognition algorithms represented by the \emph{Pure MegaLoc} baseline. 

As the reader can observe in \cref{fig:qualitative_examples}, the query images where our method outperforms the baseline are primarily those in feature-poor rural regions and highways. Such regions make single-image visual place recognition very challenging.

By taking previous query images into account and propagating its belief about the current query image's location, our method can continue to localize these images successfully.